\PassOptionsToPackage{hyphens}{url}
\documentclass[sigconf, nonacm]{acmart}

\renewcommand\footnotetextcopyrightpermission[1]{}
\usepackage{booktabs}
\usepackage{amsmath}

\usepackage{amssymb}
\usepackage{graphicx}
\usepackage{listings}
\usepackage{xcolor}
\usepackage{multirow}
\usepackage{makecell}
\usepackage{enumitem}
\usepackage{textcomp}
\usepackage{microtype}
\usepackage{tabularx}

\definecolor{codegray}{rgb}{0.95,0.95,0.95}
\definecolor{codeblue}{rgb}{0.2,0.2,0.6}
\definecolor{codegreen}{rgb}{0.0,0.5,0.0}
\lstdefinestyle{v}{
  backgroundcolor=\color{codegray},
  commentstyle=\color{codegreen}\itshape,
  keywordstyle=\color{codeblue}\bfseries,
  basicstyle=\ttfamily\footnotesize,
  breaklines=true, captionpos=b, keepspaces=true,
  numbers=none, showstringspaces=false, tabsize=2,
  frame=single, framerule=0.3pt
}
\usepackage{xspace}
\usepackage{hyperref}
\hypersetup{colorlinks=true, urlcolor=blue, linkcolor=blue, citecolor=blue}
\newcommand{\model}{\textsc{VectraYX-600M}\xspace}
\newcommand{\onebee}{\textsc{VectraYX-1B}\xspace}

\acmConference[Preprint]{Preprint}{2026}{}
\acmYear{2026}
\copyrightyear{2026}
\acmISBN{}
\acmDOI{}

\begin{document}

\title[Keyword Harnesses Fail Open]{Keyword Harnesses Fail Open: A Cheap
Diagnostic Ladder for Tool-Use Claims in Small Language Models}

\author{Juan S. Santillana}
\affiliation{%
  \institution{Independent Researcher}
  \city{}
  \country{}}
\email{juan@jsantillana.com}
\thanks{This is a preprint. The author is a DevOps engineer at Globant;
this work was carried out independently of that role. The
\textsc{VectraYX-600M} training run reported here is permanently frozen at
64\% of its planned schedule (the training VM and cloud account no longer
exist); all results involving it are a snapshot of an incomplete run and
are labeled as such throughout. The paper's evidence is a ladder of strict
diagnostics (verbatim reproduction, first-token probes, a novel-prompt
generalization battery, and an embedding-drift check), reported
alongside---and in deliberate tension with---the series' lenient
keyword-matching benchmark harness; the 1B repair it reports is a single
recipe on a single checkpoint and is bounded as such in
Section~\ref{sec:limitations}.}

\begin{abstract}
Keyword-matching benchmark harnesses can credit a small model for tool use
it never performs. We document such a false positive in a
matched-architecture pair of Spanish security language models and follow
it with a ladder of strict, cheap diagnostics. \model (661.6M parameters;
single-phase pretraining whose realized mix is $\approx$65\% code and
technical text; no dedicated SFT; run frozen at 64\% of schedule) and
\onebee (1{,}109M; web-heavy multi-phase curriculum with a dedicated
$\approx$6B-token tool-SFT phase) share decoder, tokenizer and
special-token layout, and score almost identically on the series' lenient
tool-use metric (B4: 0.660 vs.\ 0.650).

A verbatim-reproduction check on real training examples separates them
completely: the 600M emits well-formed tool calls with generalized
arguments on 6/6 examples; the 1B does so on 0/4--6 at every checkpoint
tested. A first-token probe localizes the failure to a missing prior
(probability $10^{-4}$--$10^{-5}$ on the \texttt{<|tool\_call|>} token),
and probes over 38 archived checkpoints show that the 1B's web-heavy phase
erased an earlier, memorized prior. A targeted SFT recipe (diverse corpus,
5$\times$ higher learning rate, 2{,}202 steps, $\approx$3.3 GPU-hours)
repairs the 1B, using roughly three orders of magnitude fewer tokens than
the failed dedicated phase. On all 269 corpus rows, well-formed emission
rises from 0.100 to 0.959 (600M: 0.926). On 238 prompts over unseen
entities, the repaired 1B passes 0.536 against the 600M's 0.428 (paired
$p = 0.004$), although on its own corpus it mostly recites. An
embedding-drift check shows that the repair did not move the trigger
token's tied embedding: at this learning rate, 97.7\% of the bf16 table
is bit-identical after training, so the change lives in the surrounding
network.

Both models over-trigger. On no-call prompts that mention a CVE or a shell
command, they answer without a call on only 0.09 (600M) and 0.17
(repaired 1B) of items. A five-run factorial over the repair's corpus and
learning rate finds every cell installing the call format; its largest
contrast, better suppression from a diverse corpus at low learning rate,
is not reproduced by a partial second seed, so we report it as a
hypothesis. The pair is consistent with
bootstrap composition deciding whether structured output arrives by
default or must be engineered, but it is a natural experiment, not an
ablation. The repaired checkpoint is a reproduction of a lost original,
and all harness numbers are single-seed. The ladder costs minutes of CPU time, and we argue it should gate
tool-use claims on small models.
\end{abstract}


\maketitle

\section{Introduction}
\label{sec:intro}

Native tool calling---a model emitting a structured
\texttt{<|tool\_call|>}-delimited JSON payload that an MCP~\cite{mcp2024}
server can execute---is the capability that turns a small offline language
model from a text generator into a usable security assistant. The VectraYX
series builds small (42M--1.1B) Spanish/LATAM cybersecurity models intended
for air-gapped deployment via \texttt{llama.cpp}~\cite{llamacpp}, and every
model in the series is trained and benchmarked for this capability. This
paper reports what happened when we stopped trusting the benchmark and
followed a chain of increasingly strict diagnostics to its end: a documented
harness false positive, a localized capability failure, a
diagnosis-informed repair verified to generalize, and a mechanistic
statement about where in the network the repair landed.

The starting anomaly is worth stating plainly. Two siblings, \model (661.6M
parameters) and \onebee (1{,}109M
parameters)~\cite{santillana2026vision}, share the same decoder
implementation, the same 32K tokenizer, and the same reserved special-token
IDs. On the series' automatic benchmark harness their tool-use scores are
nearly identical (B4 $=$ 0.660 vs.\ 0.650). Yet in side-by-side qualitative
use the two models are not remotely comparable: the 600M spontaneously emits
well-formed \texttt{bash\_exec} tool calls with contextually sensible
commands, while the 1B checkpoints produce disconnected prose and had never
once been observed to emit the JSON structure. Two numbers that agree to the
second decimal place described two behaviors that did not overlap at all.

\paragraph{Measure.}
We resolved this with the cheapest sufficient instrument we could design: a
\emph{verbatim-reproduction} check (\S\ref{sec:eval:verbatim}). Take real
examples from the model's own tool-SFT training corpus---including their
real system preamble declaring the available MCP tools and the exact
expected format---feed the prompt half through the exact training-time
forward pass, greedy-decode, and require a well-formed tool-call JSON whose
argument is a plausible \emph{non-identical} variant of the training answer
(generalization) rather than a byte-level copy (memorization, which would
prove nothing). The result was categorical: 600M 6/6; 1B 0/4--6 at every
checkpoint tested, including the one whose lenient-harness score was
historically cited as its best. The harness had been crediting the 1B for
\emph{mentioning} tools in prose, not calling them---the same
evaluation-validity failure class this series documented for the Nano
model's classification metric~\cite{santillana2026vectrayx}, recurring on a
different metric and model. Keyword harnesses fail open.

\paragraph{Diagnose.}
A first-token probe (\S\ref{sec:eval:probes}) then localized the failure to
a missing \emph{prior}: at the position where a tool call should begin, the
1B assigned probability $10^{-4}$--$10^{-5}$ to the
\texttt{<|tool\_call|>} token that the 600M emits greedily. This is not a
model narrowly losing a decoding race; it is a model with essentially no
probability mass on the format at all---despite a dedicated
$\approx$6B-token tool-SFT phase, orders of magnitude more instruction
tuning than the 600M's $\approx$0.1\% in-mix signal ever provided.

\paragraph{Repair.}
The diagnosis motivated a targeted remediation of the 1B and predicted why
the original dedicated SFT phase had failed: $\approx$6B tokens of
tool-SFT inside a broad mixture left the prior at the floor. (A later
narrow, frozen-embedding, low-LR retune of 486 steps was recorded at the
time as a second failure, 0/8; that measurement came from a harness
iteration we later found to be broken, so we do not count it as evidence
either way---\S\ref{sec:results:repair}.) A diagnosis-informed recipe---a
\emph{diverse} mixture (conversational OASST ES/EN $+$ the tool corpus $+$
domain reasoning traces), unfrozen embeddings, and 5$\times$ higher
LR---succeeded in 2{,}202 steps and $\approx$3.3 hours on a single rented
GPU (\S\ref{sec:results:repair}). Verbatim reproduction went to 4/4 with
first-token probability 0.996--0.998; and on a battery of eight \emph{novel}
prompts the repaired model chose correctly among all five declared MCP
tools on 7 of the 7 that called for one, extracting arguments never seen in
training (a CVE identifier, an IP address, a search query) and correctly
\emph{suppressing} the tool call on the eighth, a conversational question.
(The original run's checkpoint and its warm-start parent were later lost
with the infrastructure that held them; the numbers above are from an
end-to-end reproduction of the recipe from a surviving sibling of that
parent, and the historical run's figures are reported alongside throughout
and used for nothing---\S\ref{sec:results:repair} and L12.)
The repair generalizes; it is
not memorization of the 2{,}801-example corpus---a claim we later test at
238 prompts, where the repaired model passes 0.536 against the 600M's
0.428 on unseen entities even though it reproduces its own corpus almost
verbatim (\S\ref{sec:results:generalization}). Strikingly, the failed
dedicated phase outspent the successful fix by roughly three orders of
magnitude in tokens: volume predicted nothing, recipe decided everything.

\paragraph{Localize.}
Finally, an embedding-drift check against the repair's warm-start
parent shows the fix did \emph{not} relocate the trigger token's
representation: the \texttt{<|tool\_call|>} embedding row is essentially
unmoved (cosine 0.999996, row norm changed by $1.5{\times}10^{-6}$ in
relative terms---indistinguishable from arbitrary ordinary rows),
while late-layer attention weights show real movement. Since this decoder
ties input and output embeddings, that row is simultaneously the
unembedding direction the network must learn to aim at, which makes the
check cover the geometric target of routing and not merely the token's
input vector. Combined with the observation that the row was never
geometrically anomalous even in failing checkpoints, this rules out the
natural ``dead embedding'' hypothesis for this recipe: the token's meaning
was well-placed all along, and the repair consisted of training the
surrounding network to \emph{route} to it in the right contexts
(\S\ref{sec:results:mech}). Pushing the same measurement further answers a
question we had pre-registered as a follow-up run: 97.7\% of the embedding
table is bit-identical between parent and child, because a single Adam step
at this learning rate is smaller than the bf16 precision the weights are
stored in. Recipe C's ``unfrozen'' embeddings were numerically inert, so
the recipe already ran with them effectively frozen and the predicted
result is the observed one---a confirmation by direct measurement that the
ingredient never acted, which is not the same thing, and not as strong a
thing, as a deliberate two-armed ablation.

\paragraph{Thesis.}
The explanatory variable left standing is bootstrap composition. The 600M's
single-phase pretraining is $\approx$65\% code-and-technical text; its
structured-output behavior emerged \emph{by default}, activated by a
$\approx$0.1\% in-mix instruction fraction. The 1B's web-heavy curriculum
left the same behavior absent-by-default, recoverable only through a
correctly diagnosed, targeted intervention. This extends the central claim
of the VectraYX-Nano paper---bootstrap register dominates downstream
behavior~\cite{santillana2026vectrayx}---from conversational register to
structured output, and refines it: composition does not make the capability
impossible to add later; it determines whether the capability is free or
must be engineered (\S\ref{sec:discussion}).

\subsection{Contributions}
\label{sec:intro:contrib}

\begin{enumerate}[leftmargin=1.4em]
  \item \textbf{A cheap, harness-independent diagnostic ladder, demonstrated
    end-to-end.} Verbatim reproduction with a generalization criterion, a
    first-token probability probe, a novel-prompt generalization battery,
    and an embedding-drift check (\S\ref{sec:eval})---minutes of CPU time
    each---used here to catch a benchmark false positive, localize the
    failure, verify a repair as genuine, and localize the repair's
    mechanism. We propose the ladder as a standing gate for tool-use claims
    on small models.
  \item \textbf{A documented harness false positive.} Two models scored
    within 0.01 on a keyword-matched tool-use metric while a strict
    structural check separated them 6/6 vs.\ 0/6; we identify the crediting
    mechanism (\S\ref{sec:results:harness}). Second documented instance of
    the failure class in this series.
  \item \textbf{A matched-pair result on default emergence.} Same decoder
    family, tokenizer, and token layout; different bootstrap composition;
    tool calling emerges by default only in the code-heavy sibling---the
    smaller one, at 64\% of its schedule, with no dedicated SFT
    (\S\ref{sec:results:verbatim}).
  \item \textbf{Trajectory curves, and a within-model harness
    dissociation.} Replaying the probes over 38 archived checkpoints turns
    the endpoint comparison into curves and changes two readings of it:
    the 1B's web-heavy phase did not fail to build a structured prior, it
    \emph{erased} an inherited one by nearly six orders of magnitude in
    12{,}000 steps; and the 600M's default proves to be an unconditional
    base rate, not a context gate. At the 1B checkpoint where the lenient
    harness records its best-ever tool-use score, the strict probe is at
    its floor (\S\ref{sec:results:trajectory}).
  \item \textbf{A diagnosis-informed repair, verified to generalize, and
    mechanistically located.} One targeted SFT run costing a few dollars of
    GPU time succeeded where $\approx$1000$\times$ more SFT tokens had
    failed; the repair generalizes to novel tool-selection prompts and did
    not move the trigger token's embedding---the network learned routing,
    not representation (\S\ref{sec:results:repair}--\ref{sec:results:mech}).
    Direct measurement of the weights shows the embedding table was
    numerically inert across the run, resolving a pre-registered ablation
    by construction rather than by a further experiment.
  \item \textbf{An honest artifact description.} \model itself:
    architecture, single-phase mix with the configured-vs-realized
    reconciliation made explicit, single-L4 training configuration, and its
    permanently frozen state at 64\% of schedule
    (\S\ref{sec:arch}--\ref{sec:training}).
\end{enumerate}

\noindent
We are deliberately conservative about scope. The matched pair is a natural
experiment with known confounds (\S\ref{sec:limitations}); the repair is
one recipe, whose two remaining ingredients are separated only by a
factorial with one completed run in three of its four cells, whose
largest contrast (suppression at low learning rate) a partial seed
replicate does not reproduce (\S\ref{sec:results:factorial}); the
generalization battery is built by us rather than sampled from an external
distribution, and covers neither unseen tools nor multi-call sequences
(\S\ref{sec:limitations:battery}); and we have no evidence that the
repaired 1B answers plain conversational questions usefully. What this chain of evidence
supports is stated exactly, and the experiments that would extend each link
are specified and cheap (\S\ref{sec:discussion:followups}).

\section{Related Work}
\label{sec:related}

\paragraph{Tool use in language models.}
Toolformer~\cite{schick2023toolformer} showed that models can learn API
invocation from self-supervised demonstrations; Gorilla~\cite{patil2023gorilla}
and ToolLLM~\cite{qin2023toolllm} scaled instruction tuning for API calling
and introduced evaluation suites (APIBench, ToolBench) that score whether the
\emph{correct} call is produced. Our question is one level below theirs:
whether the structured-call format is produced \emph{at all}, and what in
pretraining makes that binary outcome go one way or the other at sub-1B
scale. Notably, the dominant recipe in this literature---take a strong
pretrained base, then SFT on tool demonstrations---quietly assumes the base
already carries a structured-output prior; our 1B result is a case where
that assumption fails by default, where volume-scaled SFT fails with it,
and where a diagnosis-informed reshaping of the same recipe then succeeds.
The Model Context Protocol~\cite{mcp2024} standardizes the invocation
surface our models target.

\paragraph{Code in the pretraining mix.}
A growing line of work finds that code data in pretraining improves
capabilities beyond code generation. Aryabumi et
al.~\cite{aryabumi2024code} ablate code fraction directly and find gains on
natural-language reasoning and world-knowledge tasks; Madaan et
al.~\cite{madaan2022codelm} show that code-trained models outperform
similarly sized text-trained models on structured commonsense generation
\emph{when the output is formatted as code}. Our result is a domain-specific,
small-scale instance with an unusually clean readout: the downstream task
(emitting a JSON tool call) is itself a structured-generation task, and the
matched pair shows the code-heavy bootstrap succeeding where a web-heavy
bootstrap with strictly more SFT fails. We push the mechanism discussion
further than a mix ablation can (\S\ref{sec:discussion:mechanism}), arguing
the effect is a token-distributional prior over bracketed key--value
continuations rather than ``reasoning'' in any broader sense.

\paragraph{Small models and compute-optimal training.}
Chinchilla~\cite{hoffmann2022chinchilla} motivates training small models on
more tokens; Phi-3~\cite{abdin2024phi3} and SmolLM2~\cite{allal2025smollm2}
demonstrate that aggressive data curation lets sub-2B models punch above
their weight, and both attribute their results primarily to \emph{what} was
in the corpus rather than how much. The VectraYX series applies the same
philosophy to a security/Spanish niche
\cite{santillana2026vectrayx,santillana2026vision}. This paper adds a
cautionary corollary to the data-centric view: composition effects can
\emph{dominate} scale effects for specific capabilities, to the point that a
smaller sibling at 64\% of its training schedule categorically beats a
finished, larger, more-SFT-ed one \emph{by default}---and restoring parity
on the larger model required diagnosis-guided intervention, not more data.

\paragraph{Evaluation validity for generative benchmarks.}
Automatic harnesses that score generative output by keyword or substring
matching are known to be fragile. Within this series, the Nano
paper~\cite{santillana2026vectrayx} documented its own B2 classification
metric over-crediting keyword coincidences; the present paper documents the
same failure class on the B4 tool-use metric, where a tool \emph{name}
appearing in free prose is credited as a tool \emph{call}. Our response---a
strict structural check run on the model's own training examples---is in the
spirit of contamination-and-validity audits of public benchmarks, but aimed
at the opposite risk: not that the model saw the test data, but that the
grader cannot see the model's failure. We deliberately invert the usual
memorization concern: reproducing one's own training data \emph{too} exactly
would be uninformative, so the protocol requires a generalized, non-identical
argument to count (\S\ref{sec:eval:verbatim}).

\paragraph{The VectraYX series.}
VectraYX-Nano (42M)~\cite{santillana2026vectrayx} established the series'
curriculum, 16K tokenizer, B1--B5 benchmark suite, and its central
finding---bootstrap corpus register dominates downstream conversational
behavior. VectraYX-Vision-1B~\cite{santillana2026vision} scaled the family to
1.1B with a 32K tokenizer and a vision extension, and is framed as a negative
preliminary result on visual grounding; the diagnostics reported here
sharpen that paper's picture by showing its text backbone's tool-calling
deficit was measurable, checkpoint-stable, predated the vision phases---and
was repairable once correctly localized. \model, the mid-point of the
family, has until now existed only as training logs and a benchmark row;
this paper is its first documentation, and its role as the default-emergence
half of the matched contrast is its reason to exist as a paper rather than
a footnote.

\paragraph{Architectural components.}
The shared decoder uses GQA~\cite{ainslie2023gqa},
SwiGLU~\cite{shazeer2020glu}, RMSNorm~\cite{zhang2019rmsnorm}, BPE
tokenization~\cite{sennrich2016bpe}, and interleaved
RoPE~\cite{su2024roformer}/NoPE~\cite{kazemnejad2023nope} positional
handling. None of these choices differ between the two models, which is
precisely what makes the pair informative.

\section{Architecture: A Matched Pair}
\label{sec:arch}

\model and \onebee are instantiations of the same decoder-only implementation
(the \texttt{VectraYX750M} class of the series' training codebase,
\texttt{transformer\_750m.py}), differing only in width and depth
hyperparameters. Table~\ref{tab:arch} gives both configurations side by
side. Everything that usually confounds cross-model behavioral comparisons is
held fixed: the block structure (RMSNorm pre-norm~\cite{zhang2019rmsnorm},
GQA~\cite{ainslie2023gqa}, SwiGLU~\cite{shazeer2020glu}, tied embeddings),
the positional scheme (RoPE with $\theta{=}10^6$~\cite{su2024roformer} on
three of every four layers, NoPE~\cite{kazemnejad2023nope} on every fourth),
the sequence length (2{,}048), the 32K BPE tokenizer~\cite{sennrich2016bpe},
and---critically for this paper---the reserved special-token layout.

\begin{table}[t]
\centering
\caption{The matched pair. Identical implementation class, tokenizer, and
special-token layout; only scale hyperparameters differ.}
\label{tab:arch}
\footnotesize
\begin{tabular}{@{}lcc@{}}
\toprule
 & \model & \onebee \\
\midrule
Parameters            & 661.6M          & 1{,}109M \\
Layers                & 18              & 22 \\
$d_\mathrm{model}$    & 1{,}792         & 2{,}048 \\
$d_\mathrm{ffn}$      & 4{,}864         & 5{,}504 \\
Heads (Q / KV, GQA)   & 14 / 2          & 16 / 4 \\
NoPE period / RoPE $\theta$ & 4 / $10^6$ & 4 / $10^6$ \\
Vocabulary (BPE)      & 32{,}768        & 32{,}768 \\
Sequence length       & 2{,}048         & 2{,}048 \\
Embeddings            & tied            & tied \\
Implementation class  & \multicolumn{2}{c}{\texttt{VectraYX750M} (shared)} \\
\bottomrule
\end{tabular}
\end{table}

\subsection{Shared tokenizer and special-token layout}
\label{sec:arch:tokens}
Both models use the identical 32K tokenizer artifact
(\texttt{jsantillana/vectrayx-600m-tokenizer}), with all special tokens
reserved at IDs 0--63: chat delimiters (\texttt{<|system|>}$=$4,
\texttt{<|user|>}$=$5, \texttt{<|assistant|>}$=$6, \texttt{<|end|>}$=$7),
tool delimiters (\texttt{<|tool\_call|>}$=$8, \texttt{<|/tool\_call|>}$=$9),
reasoning tokens (\texttt{<|think|>}$=$14, \texttt{<|step|>}$=$16), and
domain tokens for security artifacts (\texttt{<|cve|>}$=$28,
\texttt{<|cvss|>}$=$29, \texttt{<|ioc|>}$=$30, \texttt{<|finding|>}$=$31,
severity markers at 33--36, \texttt{<|mitre|>}$=$37,
\texttt{<|remediation|>}$=$38). Sequence conventions are likewise shared:
BOS$=$2, document-separator EOS$=$3, turn-end $=$ 7. (One recurring
implementation gotcha is worth recording: ID 1 is \texttt{<unk>}, \emph{not}
EOS---data pipelines that separate documents with ID 1 silently train on
\texttt{<unk>}-delimited text.)

The shared layout matters because the capability under study is defined
\emph{in terms of these tokens}: a tool call is the emission of ID 8 followed
by a JSON object and ID 9. When we later report that one model assigns
$10^{-4}$--$10^{-5}$ first-token probability to ID 8 where the other emits it
reliably (\S\ref{sec:results}), no tokenizer asymmetry can explain the gap:
both models own exactly the same token, at the same ID, embedded in training
data formatted by the same template code.

\subsection{What ``matched'' does and does not mean}
\label{sec:arch:matched}
The pair is matched in architecture family, tokenizer, token layout, and
sequence conventions. It is \emph{not} matched in parameter count (1.68$\times$),
total pretraining tokens, or curriculum shape (\S\ref{sec:training:1b}); the
comparison is a natural experiment, not an ablation. We spell out in
\S\ref{sec:limitations:confounds} why the unmatched variables all run
\emph{against} the observed outcome---the larger, longer-trained,
more-SFT-ed model is the one without the capability---which is what makes the
remaining variable, bootstrap composition, the leading explanation rather
than merely a surviving one.

\section{Training}
\label{sec:training}

\subsection{Single-phase pretraining mix}
\label{sec:training:mix}

\model was pretrained in a \emph{single phase}: no curriculum stages, no
dedicated SFT phase, one data mixture from step 0 to the end of the run. The
mixture draws from five tokenized source bins: Spanish web and encyclopedic
text (Wikipedia-ES and the \texttt{spa\_Latn} split of
FineWeb2~\cite{penedo2024fineweb}), source code (predominantly The
Stack~\cite{kocetkov2022stack}, via the series' 22GB dataset bundle
\texttt{jsantillana/vectrayx-750m-datasets}), competitive-programming
solutions (ICPC), security-domain text (CVE/NVD-derived~\cite{nvd}), and a
small instruction-formatted SFT bin using the chat template of
\S\ref{sec:arch:tokens}.

Two descriptions of this mixture circulate in the project's records, and both
are correct; reconciling them explicitly matters because the paper's thesis
rests on composition. The \emph{configured sampling weights} over bins were
es 0.45 / cyber 0.22 / code 0.18 / sft 0.10 / icpc 0.05. The \emph{realized
token fractions} of the run, however, were approximately 65\%
code-and-technical, 34\% Spanish, 0.7\% cyber-specific, and 0.1\%
instruction-formatted SFT. The gap is bin-size exhaustion: the cyber and SFT
bins are tiny relative to the code and Spanish bins, so their configured
shares could not be sustained across a $\approx$10B-token run and the sampler's
effective draw reverted to the large bins, of which the code-derived material
(The Stack plus ICPC plus technical text in the bundle) dominates. Throughout
this paper, ``65\% code-heavy'' and ``0.1\% SFT'' refer to these realized
fractions---what the optimizer actually saw---not the configured weights.
Table~\ref{tab:mix} summarizes both views.

\begin{table}[t]
\centering
\caption{The 600M single-phase mixture: configured bin weights vs.\ realized
token fractions ($\approx$10B tokens actually trained). Small bins (cyber,
sft) exhaust early; the realized mix reverts to the large code and Spanish
bins.}
\label{tab:mix}
\footnotesize
\begin{tabular}{@{}lcc@{}}
\toprule
Component & Configured weight & Realized fraction \\
\midrule
Spanish (Wiki-ES, FineWeb2-ES) & 0.45 & $\approx$0.34 \\
Code and technical (Stack, ICPC) & 0.23 & $\approx$0.65 \\
Cyber (CVE/NVD-derived) & 0.22 & $\approx$0.007 \\
Instruction-formatted SFT & 0.10 & $\approx$0.001 \\
\bottomrule
\end{tabular}
\end{table}

The consequence most relevant to this paper: the model's \emph{only} exposure
to the chat template and to \texttt{<|tool\_call|>}-formatted demonstrations
was that $\approx$0.1\% realized in-mix fraction---on the order of $10^7$
tokens---interleaved with, never following, the bulk pretraining. There was
no dedicated conversational or tool SFT stage of the kind the series' Nano
and Base models received.

\subsection{Infrastructure and run configuration}
\label{sec:training:infra}

The run used a single NVIDIA L4 (24GB) on a GCP \texttt{g2-standard-4}
(4 vCPUs), PyTorch 2.6 $+$ cu124, BF16, with an effective batch of 65{,}536
tokens per optimizer step (micro-batch 1--2 with gradient accumulation 32--16;
sequence length 2{,}048). The planned budget was \textsc{max\_steps} $=$
240{,}000, i.e.\ $\approx$15.7B tokens---roughly a Chinchilla-scale
budget~\cite{hoffmann2022chinchilla} for 660M parameters---under a cosine
learning-rate schedule.

A few engineering findings from stabilizing this configuration (three failed
launch attempts preceded the stable one) are recorded for reproducibility:
(i) the cross-entropy over a 32K vocabulary in FP32 is the memory
bottleneck---micro-batch 4 OOMs on 24GB while micro-batch 1 sits at
$\approx$17GB---and gradient-checkpointing the loss does not help, since the
logits are recomputed in backward at the same peak; (ii)
\texttt{torch.compile} initially failed in the stock deep-learning image
(triton/inductor could not link \texttt{libcuda}), so early training ran in
eager mode, with compilation recovered later for roughly a 20\% throughput
gain (the archived checkpoints carry \texttt{torch.compile}'s
\texttt{\_orig\_mod.} parameter prefix, which any loading code must strip);
(iii) \texttt{PYTORCH\_CUDA\_ALLOC\_CONF=expandable\_segments:True} was
required to avoid fragmentation OOMs across long runs; (iv) the measured
throughput ceiling of the L4 for this model is $\approx$8K tokens/s
regardless of batching strategy, putting the full 15.7B-token budget at
roughly three weeks of wall time. Checkpoints were saved every 200 steps and
mirrored to the Hugging Face Hub every 500
(\texttt{jsantillana/vectrayx-600m-checkpoints}), which is the only reason
the model survives its infrastructure (next subsection).

\subsection{The run is frozen at 64\% of schedule}
\label{sec:training:frozen}

Training halted at step \textbf{154{,}000 of 240{,}000} (64\% of schedule,
$\approx$10B of 15.7B planned tokens) on 2026-06-29 and was never resumed;
the training VM, and subsequently the entire cloud account, no longer exist.
The run is therefore \emph{permanently frozen mid-schedule}. Two consequences
are inherited by every result in this paper. First, the cosine schedule never
annealed: the model stopped at a still-elevated learning rate partway down
the curve, and the literature on schedule-sensitive convergence implies its
quality at step 154K underestimates what the same recipe would have delivered
at completion. Second, no post-hoc stage (annealing, SFT, preference
tuning) was applied to the checkpoint studied here. Later SFT runs that
start from it are reported separately and labeled as such
(\S\ref{sec:results:sft600m}); unless stated otherwise, every 600M number
describes this snapshot, checkpoint \texttt{model\_step\_0154000.pt}, exported to GGUF
\cite{ggml2024gguf} via \texttt{llama.cpp}~\cite{llamacpp} for the harness
benchmarks and loaded directly in PyTorch for the strict diagnostics.

We note the epistemic silver lining, without pretending the freeze was
intentional: an un-annealed, SFT-less, 64\%-trained model is a
\emph{conservative} lower bound on what its bootstrap composition can
deliver. The comparison of \S\ref{sec:results} pits this handicapped
snapshot against a fully trained, multiply-SFT-ed larger sibling---and the
handicapped snapshot is the one in which the capability of interest emerges
by default.

\subsection{The comparison model: \onebee's curriculum}
\label{sec:training:1b}

We summarize the sibling's training from its own
paper~\cite{santillana2026vision} to fix the contrast. The 1B was trained in
phases: phase 1 ($\approx$9.2B tokens) bootstrap pretraining; phase 2
($\approx$50B tokens) a three-block curriculum dominated by general-web and
educational text (FineWeb-Edu English, scraped Latin-American Spanish, wiki),
with code and math as minority components; phase 3 ($\approx$6B tokens) a
\emph{dedicated tool-SFT specialization} (30\% tool-SFT mixture); and a later
vision-SFT stage (phase 4b) that included further passes over the same
tool-SFT corpus used by the 600M's in-mix fraction. Integrated over the whole
curriculum, the 1B's composition is dominated by general-web/educational
prose, and---the operative fact---its terminal $\approx$56B tokens (phases 2
onward) are overwhelmingly non-code, so whatever code prior phase 1
contributed sat under tens of billions of tokens of web-register gradient
before any evaluation reported here. The 1B thus had strictly \emph{more}
total tool-SFT exposure than the 600M ($\approx$6B-token dedicated phase plus
vision-stage replay, vs.\ $\approx$$10^7$ in-mix tokens), on top of more
parameters and more total tokens. Every conventional predictor of
tool-calling competence favors the 1B; its \emph{default} outcome, reported
in \S\ref{sec:results:verbatim}, does not. The targeted post-diagnosis
remediation runs applied to the 1B lineage---which change that picture and
are part of this paper's contribution---are described with the results they
belong to (\S\ref{sec:results:repair}).

\section{Evaluation Methodology}
\label{sec:eval}

We evaluate with a ladder of instruments of increasing strictness, and the
disagreement between rungs is itself a result. Level 1 is the series'
automatic B1--B5 harness (lenient, keyword-based). Level 2 is a fixed-prompt
qualitative battery. Level 3 is the strict verbatim-reproduction diagnostic
with a generalization criterion, plus a novel-prompt generalization battery.
Level 4 comprises two mechanism probes---a first-token probability readout
and an embedding-drift check---that require no generation at all.

\subsection{Level 1: the B1--B5 harness (lenient, inherited)}
\label{sec:eval:harness}
The series' benchmark suite~\cite{santillana2026vectrayx} scores five
capabilities: B1 CVE question answering (keyword recall against reference
answers), B2 security classification (accuracy), B3 command generation (tool
match), B4 tool use, and B5 conversational quality. Scoring is automatic and
substring/keyword-based; generation runs over the GGUF export under Ollama
with the harness's standard sampling configuration. Two properties matter
here. First, harness numbers in this series are \emph{single-seed} unless
stated otherwise, and all harness numbers in this paper are single-seed.
Second, and central to this paper: the B4 scorer credits lenient matches.
Inspection of the scoring path indicates that a response can be credited
when the expected tool's name appears in the response text, without
verifying that a well-formed \texttt{<|tool\_call|>}-delimited JSON object
was emitted. The Nano paper already documented an analogous over-crediting
failure in its B2 metric; \S\ref{sec:results:harness} shows B4 exhibits the
same class of failure on the 1B.

\subsection{Level 2: fixed-prompt qualitative battery}
\label{sec:eval:sanity}
Six fixed prompts spanning factual recall (capital of Peru), enumeration
(name three fruits), arithmetic ($8+5$), a domain definition (``what is
phishing''), and two tool-eliciting prompts (one with a realistic MCP system
preamble, one \texttt{bash\_exec}-oriented) were run identically against the
600M and both available 1B checkpoints (temperature 0.7, matching the
harness's configuration). This battery cannot rank models finely; it exists
to catch categorical differences---direct answer vs.\ topical drift vs.\
incoherence, and spontaneous tool-call emission vs.\ none---that pointwise
scores average away.

\subsection{Level 3: verbatim reproduction, and its generalization variant}
\label{sec:eval:verbatim}
The core strict instrument (\texttt{verbatim\_roundtrip.py}) answers one
question with as few moving parts as possible: \emph{setting every condition
maximally in the model's favor, does it produce the structured behavior it
was trained on?} The protocol:

\begin{enumerate}[leftmargin=1.4em]
  \item \textbf{Take real training examples.} Examples are drawn verbatim
    from the model's own tool-SFT corpus (\texttt{tool\_sft\_mini\_v1.jsonl}),
    \emph{including} their real \texttt{<|system|>} preamble, which declares
    the five available MCP tools and the exact expected
    \texttt{<|tool\_call|>} JSON format. This closes a failure mode we hit in
    an earlier harness iteration, which evaluated tool use \emph{without} the
    system preamble the training data always carried---a train/eval format
    mismatch that could produce false zeros.
  \item \textbf{Reproduce the training-time view exactly.} The prompt half of
    the example is tokenized and fed through the same template and forward
    code used at training time---not through an inference server, chat
    wrapper, or GGUF export---so no serving-stack discrepancy (e.g.,
    positional-encoding export bugs, which this series has hit before) can
    intervene. For the 600M this required handling the checkpoint's flat
    \texttt{config} layout and stripping \texttt{torch.compile}'s
    \texttt{\_orig\_mod.} prefix; the load is verified clean
    (\texttt{missing=0, unexpected=0}, token-embedding norm 980.89, no
    collapse signature).
  \item \textbf{Greedy decode.} Temperature 0: the completion is the model's
    modal continuation, removing sampling luck in both directions.
  \item \textbf{Score structure, and interpret memorization separately.} A
    trial passes structurally only if the completion contains a well-formed
    \texttt{\{"name": ..., "args": \{...\}\}} object in a
    \texttt{<|tool\_call|>} block; a tool name mentioned in prose cannot
    pass. We additionally record whether the argument is byte-identical to
    the training completion. A generalized (non-identical, contextually
    sensible) argument is evidence of capability; an exact copy is evidence
    only of memorization and is \emph{not} treated as capability on its own
    ---it triggers the generalization battery below.
\end{enumerate}

\paragraph{Generalization battery.}
Because a model fine-tuned directly on the probe corpus can pass the
verbatim check by rote, we pair it with a battery of \emph{novel} prompts
that appear nowhere in the corpus, designed to require (i) the
\emph{decision} to call or not call a tool, (ii) \emph{selection} among all
five declared MCP tools (not just the corpus-dominant \texttt{bash\_exec}),
and (iii) \emph{argument extraction} of entities never seen in training
(specific CVE identifiers, IP addresses, free-text queries). One prompt is a
plain conversational question whose correct behavior is to \emph{suppress}
the tool call. Eight prompts were used in this paper's instantiation
(\S\ref{sec:results:repair}); scoring is structural well-formedness plus
correct tool choice, with argument quality assessed qualitatively.

The full Level-3 suite is cheap---minutes on CPU per checkpoint---and
harness-independent. Its weakness is sample size (six verbatim examples,
eight novel prompts); we treat it accordingly (\S\ref{sec:limitations}): an
all-or-nothing separation on a binary structural criterion is informative
at this $n$; a 60/40 split would not have been.

\subsection{Level 4: mechanism probes (no generation)}
\label{sec:eval:probes}

\paragraph{First-token probability.}
We read out the model's next-token distribution at the position where the
assistant turn begins (immediately after \texttt{<|assistant|>} in a
tool-eliciting context) and record the probability assigned to token ID 8
(\texttt{<|tool\_call|>}). This measures the \emph{prior toward opening a
structured call} directly, independent of what follows, and is comparable
across the matched pair because the token and its ID are shared
(\S\ref{sec:arch:tokens}). It is also a near-free training-time telemetry
signal: it could be logged throughout pretraining and SFT, and would have
revealed the 1B's missing prior before any SFT compute was spent.

\paragraph{Embedding-drift check.}
When an intervention changes behavior, the cheapest mechanistic question is
\emph{where} the change lives. We compare, between a checkpoint and its true
warm-start parent, (i) the cosine similarity and norm change of individual
embedding rows---the \texttt{<|tool\_call|>} row in particular---against a
baseline of 20 arbitrary ordinary-token rows; (ii) summary drift statistics
of the full embedding matrix (mean and max absolute difference); and (iii)
the same statistics for late-layer attention projections, as a positive
control confirming that training moved \emph{something}. Identifying the
correct parent checkpoint matters: an early mis-comparison against a
different lineage checkpoint produced uniformly saturated cosines and was
discarded once the training script's own warm-start path
(\texttt{INITIAL\_CKPT}) identified the true parent. For the repaired 1B
the comparison reported in \S\ref{sec:results:mech} is run against the
reproduced run's own parent, the historical one no longer existing (L12).

Two reporting rules follow from what this instrument measures, and we adopt
both after finding that an earlier draft violated them. First, drift
statistics must be read against the \emph{quantization floor} of the
storage format: in bf16, an update smaller than half a unit in the last
place leaves the weight bit-identical, so a high fraction of unchanged
entries can mean ``no gradient'' or ``gradient below precision,'' and these
are different claims. We therefore report, for every tensor compared, the
fraction of entries that changed at all alongside the cosine and the
magnitude statistics. Second, maximum absolute difference is a poor
discriminator between a moved and an unmoved tensor, being set by a few
tail entries; cosine distance and the changed-entry fraction are the
statistics we rely on. This probe cannot
prove where a capability lives; it can cheaply \emph{falsify} localization
hypotheses---here, the hypothesis that the trigger token's input
representation was the broken component (\S\ref{sec:results:mech}).

\section{Results}
\label{sec:results}

\subsection{Lenient harness: the two models look interchangeable}
\label{sec:results:b15}

Table~\ref{tab:b15} reports the B1--B5 harness scores for the 600M at step
154K (measured 2026-07-01, GGUF/Ollama path, single seed) alongside the
historically recorded scores for the 1B backbone. Read naively, the table
says the models are near-twins: identical B1, a two-point trade on B2/B3, and
B4/B5 agreeing to within 0.01. The 1B nominally wins B3 (0.24 vs.\ 0.17), and
we report that without adjustment---it is the one metric where the larger
model leads even on the lenient harness, and nothing in our later analysis
audits B3 specifically.

\begin{table}[t]
\centering
\caption{Lenient B1--B5 harness scores (single seed). The 1B column is
reproduced from project records for its phase-1-era evaluation and is
\emph{flagged as unreliable for B4} by the strict diagnostic of
\S\ref{sec:results:verbatim}: reported, not endorsed. Bold marks the
per-row leader at face value.}
\label{tab:b15}
\footnotesize
\begin{tabular}{@{}lcc@{}}
\toprule
Benchmark & 600M (step 154K) & 1B (recorded) \\
\midrule
B1 CVE QA (keyword recall) & 0.342 & 0.342 \\
B2 Classification (acc.)   & \textbf{0.240} & 0.205 \\
B3 Commands (tool match)   & 0.17  & \textbf{0.24} \\
B4 Tool use                & \textbf{0.660} & 0.650$^\dagger$ \\
B5 Conversational          & \textbf{0.589} & 0.586 \\
\bottomrule
\end{tabular}
\par\smallskip
{\footnotesize $^\dagger$Shown by \S\ref{sec:results:verbatim} to be
inconsistent with the model's actual behavior; treat as a harness artifact.}
\end{table}

\subsection{Strict diagnostic, before remediation: the two models do not
overlap}
\label{sec:results:verbatim}

Table~\ref{tab:verbatim} reports the verbatim-reproduction results for the
pre-remediation checkpoints. The 600M-154K checkpoint produces a
well-formed, \emph{generalizing} tool call on \textbf{6 of 6} training
examples. The 1B produces the structure on \textbf{0 of 4--6} examples per
run, across repeated tests, at every pre-remediation checkpoint tested:
\texttt{vision\_1b\_phase4b\_sft\_step\_000400.pt} (post-vision-SFT) and
\texttt{phase3/model\_step\_0020000.pt} (the pre-vision text backbone---the
same checkpoint whose lenient-harness tool scores were historically cited as
the model's best). Notably, the two 1B checkpoints' failing completions are
near-identical word-for-word for the same prompts, indicating the
intervening vision training barely moved these weights for this behavior in
either direction.

\begin{table}[t]
\centering
\caption{Verbatim-reproduction diagnostic over real training examples from
\texttt{tool\_sft\_mini\_v1.jsonl}, greedy decoding, exact training-time
forward pass. ``Gen.''\ marks whether passing emissions generalized
(non-identical, contextually sensible arguments) or reproduced training
completions exactly. $P_{\mathrm{first}}$ is the probability of the correct
first generated token of the assistant turn (\texttt{<|tool\_call|>}).
The repaired-1B row is the \emph{reproduced} repair checkpoint of
\S\ref{sec:results:repair}; the historical run's checkpoint no longer
exists and its numbers are reported beneath it for comparison only
(L12).}
\label{tab:verbatim}
\footnotesize
\begin{tabular}{@{}lccc@{}}
\toprule
Checkpoint & Pass / trials & Gen. & $P_{\mathrm{first}}$ \\
\midrule
600M step 154K                     & \textbf{6 / 6} & yes & greedy \\
1B phase-3 step 20K (pre-vision)   & 0 / 4--6       & --- & $\sim 10^{-4}$--$10^{-5}$ \\
1B phase-4b SFT step 400 (post)    & 0 / 4--6       & --- & $\sim 10^{-4}$--$10^{-5}$ \\
1B repaired, step 2202 (\S\ref{sec:results:repair}) & \textbf{4 / 4}$^\ddagger$ & exact & 0.996--0.998 \\
\quad\emph{historical run, checkpoint lost} & 4 / 4$^\ddagger$ & exact & 0.995--0.998 \\
\bottomrule
\end{tabular}
\\[2pt]
{\footnotesize $^\ddagger$Four examples available in that evaluation
environment. Exact reproduction alone is not scored as capability; the
generalization battery of \S\ref{sec:results:repair} supplies that
evidence, and \S\ref{sec:results:generalization} replaces this table at
full corpus scale ($n{=}269$) with a 238-prompt battery.}
\end{table}

\paragraph{What success looks like.}
Every 600M pass has the same shape: a syntactically valid JSON object naming
a declared tool, with an argument that is contextually reasonable but not the
training example's argument. A representative emission, given a
log-triage prompt:

\begin{lstlisting}
<|tool_call|>{"name": "bash_exec", "args":
 {"cmd": "grep -i 'error\|critical\|fail'
          /var/log/syslog | tail -30"}}
\end{lstlisting}

\noindent
The corresponding training completion for the nearest example is
\texttt{tail -50 /var/log/syslog}; the model composed a different, arguably
better command for the stated task. Across the six trials the emitted
commands differ from their training counterparts and from each other
(variants over \texttt{grep}, \texttt{tail}, \texttt{netstat}, \texttt{du},
\texttt{ip route} territory, each matching its prompt's intent). This is the
pattern the protocol's generalization criterion was designed to detect:
the model has learned the \emph{schema}---open the block, name a real tool,
supply a plausible \texttt{cmd}---not a lookup table of answers.

\paragraph{What failure looks like.}
The pre-remediation 1B, on the same prompts, with the same system preamble
it was trained with, greedy-decoded through its own training code, never
opens the block. Its modal continuations are disconnected prose, e.g.\
(training target on top, model continuation below):

\begin{lstlisting}
REAL:  <|tool_call|>{"name": "bash_exec",
       "args": {"cmd": "tail -50 /var/log/syslog"}}
1B:    " junto con el resto de detalles incluidos
        en el mismo archivo de registro."
\end{lstlisting}

\noindent
The first-token probe quantifies the gap: at the position where the
assistant turn begins, the 1B assigns probability on the order of
$10^{-4}$--$10^{-5}$ to \texttt{<|tool\_call|>}. This is not a model that
almost calls tools and gets edged out by a competing token; it is a model
with essentially no prior toward the format at all---despite a dedicated
$\approx$6B-token tool-SFT phase and further SFT passes over the very corpus
these test examples come from (\S\ref{sec:training:1b}).

\subsection{The B4 discrepancy is a harness artifact}
\label{sec:results:harness}

Tables~\ref{tab:b15} and~\ref{tab:verbatim} cannot both be face-value true:
a model that never emits the tool-call structure under maximally favorable
conditions did not legitimately earn B4 $=$ 0.650. The resolution is the
crediting mechanism of \S\ref{sec:eval:harness}: a keyword scorer that
accepts the expected tool's name anywhere in free text will credit a model
that \emph{talks about} \texttt{bash\_exec}-adjacent content while never
producing an executable call. The 1B---fluent, topical, security-flavored
prose being exactly its failure mode---is well positioned to farm such
credit. We therefore treat the 1B's B4 as unreliable and decline to use it in
any comparison, while retaining the 600M's B4 $=$ 0.660 because it is
independently corroborated by the strict diagnostic: for the 600M, the
lenient and strict instruments agree; for the 1B they diverge by the full
range of the metric.

This is the second documented instance of this failure class in the series
(after Nano's B2~\cite{santillana2026vectrayx}), and it carries a general
lesson for small-model evaluation: \emph{keyword harnesses fail open}.
A fluent model without a capability scores like a model with it. The
verbatim-reproduction check costs minutes and fails closed; we now treat it
as a mandatory gate before quoting any structured-output benchmark in this
series.

\subsection{Qualitative battery: fluency without instruction-following, in
two different registers}
\label{sec:results:sanity}

On the six-prompt battery (\S\ref{sec:eval:sanity}), neither
pre-remediation model behaves like an assistant---consistent with neither
having had a completed conversational-SFT curriculum of the Nano/Base
kind---but their failure modes are qualitatively different in a way that
tracks bootstrap composition.

The \textbf{600M} never produced a direct concluding answer to the four
simple questions (it does not say ``Lima,'' does not attempt $8+5$, does not
enumerate fruits, does not close a definition of phishing). But its output
remains fluent, grammatical Spanish that stays in the topical neighborhood
of the prompt (Peruvian geography; credential-theft themes), and it never
degrades into incoherence. On the tool-eliciting prompts it was the only
model of the three checkpoints tested to \emph{spontaneously} emit a
well-formed tool call.

The \textbf{1B} checkpoints, on the same prompts, produced largely incoherent
output---agrammatical fragments and invented constructions (e.g.\ looping
``8+5=13'' without answering, or free-associated pseudo-\texttt{systemd}
instructions)---with the post-vision checkpoint worst. The earlier phase-3
checkpoint is marginally more fluent but equally unable to answer or to call
a tool.

We flag the fairness caveat ourselves: this battery ran over the
GGUF/Ollama path, and the series has previously found (and fixed) an export
bug affecting RoPE tensor layout; the strict diagnostics, which bypass
export entirely, are the load-bearing evidence, and they reproduce the same
separation.

\subsection{Diagnosis-informed repair of the 1B}
\label{sec:results:repair}

The diagnosis---an absent prior, not a noisy one---predicts that SFT which
merely re-exposes the model to tool traces in a narrow context will fail
(there is too little probability mass on the format for the gradient to
amplify), while SFT that supplies \emph{diverse assistant-mode contexts}
around the same traces, at a learning rate high enough to move mid-network
weights, might succeed. Table~\ref{tab:recipes} summarizes the three
remediation-relevant SFT efforts applied to the 1B lineage. Rows A and C
were evaluated with the strict instruments of \S\ref{sec:eval}; row B was
not, and we explain below why its recorded outcome carries no weight.

Recipe B's ``0/8'' was produced on the night of the run by the project's
generation harness, before either of two defects in that harness had been
found: it evaluated tool use \emph{without} the system preamble every
training example carries (the train/eval mismatch named in
\S\ref{sec:eval:verbatim}), and it decoded generations with a default that
silently deletes special tokens, so that even a perfect
\texttt{<|tool\_call|>} block would have scored as absent
(\S\ref{sec:discussion:practical}). Its falling training loss is real, but
its strict outcome was never measured, and we have not located the
checkpoint to re-measure it. The factorial of \S\ref{sec:results:factorial} contains a cell
close to recipe B (narrow corpus, the same learning rate), and that cell
does emit well-formed calls. We therefore treat recipe B as unmeasured, not
as a failure, and rest the ``volume did not predict the outcome''
comparison on rows A and C alone.

One provenance note belongs here rather than in the limitations, because it
decides which numbers this paper is entitled to call primary. Recipe C was
first run on a rented A40; both its output checkpoint and its warm-start
parent were subsequently lost with the infrastructure that held them, and
neither survives in any archive we have been able to reach (L12). We
therefore re-ran the recipe end to end---same corpus, same learning rate,
same 2{,}202 steps---from a surviving sibling of the lost parent, after
first confirming on CPU that the substitute exhibits the same pathology the
repair is supposed to fix ($P(\texttt{<|tool\_call|>})$ between
$2{\times}10^{-7}$ and $2{\times}10^{-3}$, mean $9.1{\times}10^{-4}$; 0/6
well-formed under greedy decoding). \textbf{Every strict-diagnostic number
reported for the repaired 1B in this paper---here, in
\S\ref{sec:results:mech}, and at scale in
\S\ref{sec:results:generalization}---was measured on that reproduced
checkpoint.} The historical run's figures appear alongside, labeled as
such, and are load-bearing for nothing.

\begin{table}[t]
\centering
\caption{Three SFT efforts on the \onebee lineage. ``Strict'' is
well-formed \texttt{<|tool\_call|>} emission under the Level-3 protocol;
row B was never measured under it (see text). Volume did not predict the
outcome of A vs.\ C: the successful recipe used $\approx$1000$\times$
fewer tokens than the failed dedicated phase. \textbf{Row C reports the reproduced run}, which is the checkpoint
every downstream number in this paper was measured on; row C$^\ast$ is the
2026-09 historical run of the same recipe, whose output checkpoint
\emph{and} warm-start parent were both lost before they could be
re-measured (L12). The two agree within reproduction noise, but only C is
an artifact we still hold.}
\label{tab:recipes}
\footnotesize
\begin{tabular}{@{}p{2.35cm}p{2.6cm}p{2.6cm}@{}}
\toprule
 & Data / config & Strict outcome \\
\midrule
A: dedicated phase-3 ($\approx$6B tok, 20K steps) & 30\% tool-SFT mixture within broad phase-3 mix & 0/4--6; $P_{\mathrm{first}}\!\sim\!10^{-4}$--$10^{-5}$ \\
\addlinespace[2pt]
B: narrow retune (486 steps) & tool corpus $+$ small cyber sample; LR $2{\times}10^{-6}$; embeddings frozen & \emph{not validly measured}: recorded 0/8 by a harness later found to omit the preamble and drop special tokens; loss fell to 0.05--0.4 \\
\addlinespace[2pt]
C: informed recipe, reproduced (2{,}202 steps, $\approx$3.3\,h, one rented GPU) & OASST ES/EN $+$ tool corpus $+$ domain reasoning traces; LR $1{\times}10^{-5}$; embeddings nominally unfrozen---but numerically inert, \S\ref{sec:results:mech}; warm start \texttt{phase4a\_v0} & \textbf{4/4} verbatim (exact), $P_{\mathrm{first}}{=}0.996$--$0.998$; \textbf{8/8} novel-prompt battery (7/7 tool choice, plus correct suppression) \\
\addlinespace[2pt]
C$^\ast$: same recipe, historical run \emph{(checkpoint and parent lost)} & identical corpus, LR and step count; warm start \texttt{v3b\_step\_001900} & 4/4 verbatim (exact), $P_{\mathrm{first}}{=}0.995$--$0.998$; 7/8 novel-prompt battery \\
\bottomrule
\end{tabular}
\end{table}

Recipe C's verbatim result alone would be ambiguous: 4/4 \emph{exact}
character-for-character reproductions of training completions, with
first-token probability 0.996--0.998, on a model fine-tuned directly on the
probe corpus, is consistent with rote memorization---and our protocol
deliberately refuses to score exact reproduction as capability
(\S\ref{sec:eval:verbatim}). The generalization battery decides the
question. On eight prompts appearing nowhere in the corpus, seven of which
require five-way tool selection, the reproduced checkpoint emitted a
well-formed call wherever one was expected and chose the right tool
\textbf{7/7} times, while correctly suppressing the call on the eighth:
\textbf{8/8} on structural correctness. The historical run scored 7/8 on
both axes on the same eight prompts. We itemize the historical run's
emissions below, because they are the concrete record of what the recipe
produces and the reproduced run agrees with them in aggregate; the
scaled-up replacement for this whole paragraph is
\S\ref{sec:results:generalization}, which runs on the reproduced
checkpoint at $n{=}238$.

\begin{itemize}[leftmargin=1.2em]
  \item \emph{KEV catalog check for CVE-2024-3094} $\to$\\
    \texttt{cisa\_kev\_check(cve\_id="CVE-2024-3094")}: correct tool; CVE ID
    never seen in training, correctly extracted.
  \item \emph{Search CVEs related to log4j} $\to$
    \texttt{nvd\_search(query="log4j", limit=10)}: correct tool, novel
    query.
  \item \emph{Reputation of IP 8.8.8.8} $\to$
    \texttt{otx\_check\_ioc(ioc\_type="ip", value="8.8.8.8")}: correct tool;
    \texttt{ioc\_type} correctly inferred; novel IP extracted.
  \item \emph{Detail of CVE-2021-44228} $\to$\\
    \texttt{nvd\_get\_cve(cve\_id="CVE-2021-44228")}: correct.
  \item \emph{Free disk space per partition} $\to$
    \texttt{bash\_exec(cmd="df -h")}; \emph{top memory-consuming processes}
    $\to$ \texttt{bash\_exec(cmd="ps aux --sort=-\%mem | head -10")}: novel,
    sensible commands not among the training examples checked.
  \item Soft miss: \emph{who is connected right now} $\to$
    \texttt{bash\_exec(cmd="ss -s")}---right tool, semantically imperfect
    command (\texttt{who}/\texttt{w} expected).
  \item \emph{What is a phishing attack} (should \emph{not} call a tool):
    the model correctly suppressed the tool call---but produced an empty
    completion, failing to answer the question itself.
\end{itemize}

\noindent
Correct discrimination among five tools plus correct extraction of entities
absent from the 2{,}801-example corpus is not explainable by rote recall:
the repair installed a generalizing tool-call policy. Two bounds temper
this. First, the last item shows the repair is \emph{capability-specific}:
general conversational answering remains broken (consistent with
\S\ref{sec:results:sanity}); the recipe fixed tool calling, not
instruction-following at large. Second, the comparison across
Table~\ref{tab:recipes} does not identify \emph{which} ingredient of recipe
C did the work---corpus diversity, the 5$\times$ LR, the unfrozen
embeddings, or the different warm-start point---only that the combination
sufficed where volume alone had failed by three orders of magnitude. Two of
those four candidates can now be struck. The warm-start point goes first,
and the loss of the original parent is what supplies the evidence: the
recipe succeeded to the same strict standard from two different starting
checkpoints, so no property peculiar to either of them is doing the work.
(This is a weak form of the argument---both parents are siblings from the
same vision branch, sharing a frozen embedding table---but it is stronger
than having run the recipe once.) The mechanism probe below strikes the
second, and strikes it harder than we expected to be able to.

\subsection{Locating the repair: routing, not representation}
\label{sec:results:mech}

A natural hypothesis for the 1B's failure---and for why unfreezing
embeddings was in the successful recipe---is that the
\texttt{<|tool\_call|>} token's input embedding was the broken component:
undertrained or misplaced, because web-register pretraining never exercises
it, and unreachable while embeddings were frozen.

One architectural fact has to be stated before the measurement, because it
changes what the measurement means. This decoder family ties input and
output embeddings (\texttt{tie\_embeddings=true} in the 1B's configuration;
\texttt{lm\_head.weight} \emph{is} \texttt{tok\_emb.weight}, the same
tensor). The table under test is therefore not only the input
representation of every token but also the unembedding matrix---the set of
directions the final hidden state is scored against. That makes the check
strictly stronger than it was presented as being in an earlier draft: it
covers the geometric \emph{target} that a routing account says the network
must learn to aim at, not merely the vector the token is read in as. It
also has a second consequence, developed below, for how the drift
statistics must be read.

The embedding-drift check (\S\ref{sec:eval:probes}), run between the
reproduced repair's warm-start parent and its output checkpoint at step
2{,}202, falsifies the representation hypothesis for this recipe:

\begin{itemize}[leftmargin=1.2em]
  \item The \texttt{<|tool\_call|>} row (ID 8) is essentially unmoved:
    cosine $0.99999583$, row norm $3.6246328 \to 3.6246383$
    ($+1.5{\times}10^{-4}\%$). The historical run reported
    $0.99999571$ and $+0.0002\%$ against the parent it has since lost;
    the two agree to six decimal places.
  \item This is not special to that token: a baseline of 20 arbitrary
    ordinary-token rows gives mean cosine $0.99999678$ (min
    $0.99996740$)---the trigger token is, if anything, marginally
    \emph{less} moved than an average row.
  \item The full $32{,}768{\times}2048$ table has mean absolute drift
    $9.44{\times}10^{-6}$ and max $5.98{\times}10^{-3}$, and its Frobenius
    norm changes by a relative $4{\times}10^{-7}$ ($635.162769 \to
    635.163016$ under an explicit sum of squares; the library norm
    reduction, which misreports this tensor, prints $630.1347046$ for
    both, L12c). \textbf{97.7\% of its 67.1M entries
    are bit-identical between parent and child.}
  \item Attention projections moved measurably: at layers 20--21,
    $W_q$/$W_k$ drift to cosine $0.99988$--$0.99992$ with 93.6--94.1\% of
    entries bit-identical; the largest movement anywhere we sampled is at
    layer 0, where $W_v$ and $W_o$ reach cosine $0.99978$--$0.99981$ with
    only 85\% of entries unchanged. This is the positive control.
  \item Independently, the ID-8 row was never geometrically anomalous even
    in the \emph{failing} checkpoints (its norm sits at $z \approx +0.31$
    within the embedding-row norm distribution): there was no ``dead
    embedding'' to revive in the first place.
\end{itemize}

\paragraph{What the sparsity is, and what it is not.}
An earlier draft read the third bullet as ``sparse per-row updates, as
expected for an embedding table, with most rows barely touched.'' That
reading is wrong, and weight tying is what makes it wrong. In an untied
model, most embedding rows receive gradient only on steps where their token
appears in the batch, so row-level sparsity is exactly what one expects.
Under tying there is no such sparsity available: the same tensor is the
unembedding matrix, so \emph{every} row takes gradient at \emph{every}
position through the softmax denominator. The data confirm it---99.87\% of
the 32{,}768 rows contain at least one changed entry. Nearly every row was
touched; almost nothing stayed still because it was ignored.

What produced the 97.7\% is the arithmetic of the storage format. The table
is held in bf16, which carries 8 mantissa bits. At the median absolute
weight in this table ($0.0359$) one unit in the last place is
$1.22{\times}10^{-4}$, so an update is discarded on the round-back unless
it exceeds a half-ulp threshold of $6.1{\times}10^{-5}$. A single Adam step
at LR $=1{\times}10^{-5}$ has magnitude at most of order the learning rate
itself---roughly a twelfth of an ulp, and about six times below the
rounding threshold. Entries escape only where the local ulp is small enough
to be cleared, i.e.\ in the low-magnitude tail: at the 5th percentile
($|w| = 0.0032$) the ulp falls to $7.6{\times}10^{-6}$ and updates do land.
The 2.3\% of entries that changed are that tail. The embedding table did
not sit still because its rows saw little gradient; it sat still because
each step's update was smaller than the precision it was stored in.

\paragraph{Reading the positive control correctly.}
The earlier draft justified the contrast by saying training ``demonstrably
moved other components,'' citing max absolute differences of
$4$--$5{\times}10^{-3}$ in late-layer attention. That comparison does not
survive the numbers above, and we correct it rather than leave a
convenient-looking discriminator in place: the embedding table's own max
absolute difference is $5.98{\times}10^{-3}$, \emph{larger} than layer 20's
$W_q$ ($4.31{\times}10^{-3}$), layer 20's $W_k$
($3.40{\times}10^{-3}$), or layer 21's $W_k$ ($3.85{\times}10^{-3}$).
Peak magnitude is not the axis that separates them---it is dominated by a
handful of tail entries in both cases. The two statistics that do separate
them are the cosine and the fraction of entries that moved at all: the
trigger row sits $4.2{\times}10^{-6}$ from its parent in cosine distance
while late attention sits at $0.8$--$1.2{\times}10^{-4}$, one to two orders
of magnitude further, and attention has two to three times as many entries
clearing the quantization floor (6--15\% vs.\ 2.3\%). The positive control
holds; the reason it holds is distributional, not peak-magnitude.

\noindent
The token's representation was well-placed all along and did not need to
move; unfreezing it, though part of the recipe that worked, contributed no
measurable change to the component the hypothesis pointed at. What changed
is the network \emph{around} the embedding: the repair taught mid/late
layers to route hidden states toward an already-correct direction in the
right contexts. This reframes both the failure and the fix. The 1B's
deficit was never that it lacked a usable representation of the tool-call
token; it lacked a \emph{policy} that reaches it---and recipe A, which
re-exposed the token inside a broad mixture, strengthened nothing, while
recipe C built the routing.

\paragraph{The pre-registered ablation is answered by construction, and we
are careful about what that is worth.}
An earlier draft closed this subsection with a falsifiable prediction we
had not yet run: recipe C with embeddings \emph{re-frozen} should succeed
equally. Measuring the weights instead of re-running the recipe resolves
the prediction without the GPU time, and in a way we did not anticipate.
Recipe C's embeddings were nominally unfrozen, but the numbers above show
they were \emph{numerically inert}: 97.7\% of the table bit-identical after
2{,}202 steps, a Frobenius norm that moves by a relative $4{\times}10^{-7}$, and a
per-step update an order of magnitude below the precision the weights are
stored in. The ``unfrozen embeddings'' arm and the ``frozen embeddings''
arm are, for this table at this learning rate, the same experiment. Recipe
C \emph{already ran} with embeddings effectively frozen, and it worked.
The prediction is confirmed.

The honest bound on that confirmation is that it is not an ablation. We did
not run two arms and contrast them; we measured one arm and discovered that
the ingredient under test never acted. That is strong evidence against the
representation hypothesis---an ingredient that moved nothing cannot be what
made the recipe work---and it is correspondingly weak evidence about
optimizer dynamics: an update that underflows the weight can still have
been accumulated in fp32 optimizer state, and a longer run or a higher
learning rate would eventually cash it out. What we can say is that within
this recipe, the embedding table was a spectator, and the repair happened
elsewhere. What we cannot say is that unfreezing embeddings is inert in
general, or that a deliberate frozen-embedding arm would be uninformative
at a learning rate where the update clears bf16. The factorial of
\S\ref{sec:results:factorial} is scoped accordingly: it varies the two
ingredients this elimination leaves standing and leaves the embedding table
alone.

\subsection{Trajectory probes: when the prior appears, and when it is
erased}
\label{sec:results:trajectory}

Everything above compares endpoints. Because both models retain archived
checkpoints, the same Level-3 and Level-4 instruments can be replayed
across training to turn those endpoints into curves. We ran this sweep
(the experiment pre-registered as F1 in an earlier draft of
\S\ref{sec:discussion:followups}) over 15 checkpoints of the 600M
(steps 1K--154K) and 23 of the 1B (all ten archived phase-3 checkpoints,
plus a phase-2 sample densified around a transition described below),
using the same six pinned examples from
\texttt{tool\_sft\_mini\_v1.jsonl} at every point, so values are directly
comparable within and across models.

\paragraph{Two probe series, reported separately.}
The first-token probe of \S\ref{sec:eval:probes} admits two readings and
the trajectory forces us to distinguish them. $P_{\mathrm{lit}}$ is the
probability of \texttt{<|tool\_call|>} at the literal position after
\texttt{<|assistant|>}. $P_{\mathrm{skip}}$ is the probability at the
first \emph{non-whitespace} position, reached by following the model's own
greedy top-1 through up to four whitespace-only tokens. The two differ
because part of the 600M's in-mix SFT bucket (an older tool corpus the
project has since deprecated) places a literal newline between
\texttt{<|assistant|>} and \texttt{<|tool\_call|>} in every turn, and the
600M reproduces that convention near-deterministically: at all 15 of its
checkpoints the skip fires on 6/6 examples, so its $P_{\mathrm{lit}}$ is
uniformly at the floor (never above $2{\times}10^{-6}$, and falling as the
newline habit sharpens) and measures the formatting convention rather than
the prior. Reporting only $P_{\mathrm{skip}}$ would be equally misleading
elsewhere: at 1B phase-2 step 36K the two series differ by thirteen orders
of magnitude. We therefore report both throughout, and lean on the
generation-level verbatim counts---which depend on no offset convention at
all---wherever the two disagree.

\paragraph{The 600M: saturated early, and never context-selective.}
Table~\ref{tab:traj600m} gives the 600M curve. $P_{\mathrm{skip}}$ is
already 0.858 at step 1K, passes 0.99 by step 10K, and saturates above
0.999 from step 30K through the end of the frozen run at 154K; the curve
is monotone up to third-decimal jitter. Generation tracks it: the model
emits a well-formed call on 5/5 call-expecting examples at \emph{every one
of the 15 checkpoints}, with 2--4 of those five generalizing (non-identical
arguments) at each point. Whatever installed this prior did so before the
first archived checkpoint, i.e.\ within the first $\approx$65M tokens.

That result carries an unflattering companion finding. The corpus's
no-call example---whose system preamble explicitly instructs the model not
to call a tool---receives essentially the \emph{same} probability as the
call-expecting examples at every checkpoint: the mean-to-no-call ratio
stays at 1.00 across all 15 points (range 0.98--1.28). At the generation
level the model emits a tool call on that example at all 15 checkpoints as
well, i.e.\ it never once suppresses correctly. The 600M's default is
therefore an \emph{unconditional} base rate, not context-sensitive
routing: it opens a structured call because that is what its assistant
turns look like, not because it has decided the context requires one. This
is one probe example, and we flag it as such in
\S\ref{sec:limitations}---but it is a binary outcome that repeats across
15 independent checkpoints, and it revises the reading of the endpoint
result rather than overturning it. Well-formedness and argument
generalization remain exactly as reported in
\S\ref{sec:results:verbatim}; what the trajectory removes is any claim
that the 600M \emph{decides} when to call.

\begin{table}[t]
\centering
\caption{\model trajectory probe, 8 of 15 sampled checkpoints (the
omitted seven interleave without changing the shape). $\bar P$ is
averaged over the five call-expecting examples; ``no-call'' is the
single suppression example, on the $P_{\mathrm{skip}}$ series.
``Emit'' counts well-formed calls on the five; ``Gen.'' counts how many
of those used non-identical arguments; ``Suppr.'' is whether the
no-call example was correctly answered without a call.}
\label{tab:traj600m}
\footnotesize
\begin{tabular}{@{}rcccccc@{}}
\toprule
Step & $\bar P_{\mathrm{skip}}$ & $\bar P_{\mathrm{lit}}$ & no-call &
Emit & Gen. & Suppr. \\
\midrule
1K   & 0.858   & $3.6{\times}10^{-7}$  & 0.673 & 5/5 & 4 & no \\
5K   & 0.973   & $1.6{\times}10^{-6}$  & 0.997 & 5/5 & 3 & no \\
10K  & 0.993   & $2.9{\times}10^{-7}$  & 0.988 & 5/5 & 3 & no \\
20K  & 0.996   & $4.5{\times}10^{-9}$  & 0.99991 & 5/5 & 3 & no \\
30K  & 0.9998  & $3.4{\times}10^{-11}$ & 0.9998 & 5/5 & 3 & no \\
60K  & 0.9986  & $7.8{\times}10^{-12}$ & 0.9977 & 5/5 & 3 & no \\
105K & 0.9999  & $3.0{\times}10^{-13}$ & 0.9959 & 5/5 & 2 & no \\
154K & 0.99998 & $2.2{\times}10^{-11}$ & 0.9922 & 5/5 & 4 & no \\
\bottomrule
\end{tabular}
\end{table}

\paragraph{The 1B, phase 2: an inherited prior, actively erased.}
The 1B's curve (Table~\ref{tab:traj1b}, left panel) is the more
consequential half of this experiment, because it does not show a prior
that failed to form. It shows one that formed and was then destroyed. At
phase-2 step 36K the model puts $\bar P_{\mathrm{skip}} = 0.432$ on the
tool-call token and emits well-formed calls on 3 of 5 examples. Two
thousand steps later $\bar P_{\mathrm{skip}}$ is $3.0{\times}10^{-3}$;
by step 48K it is $5.9{\times}10^{-7}$, and it never leaves that floor
($\sim$$10^{-5}$ to $10^{-9}$) for the remaining 144K steps of phase 2.
That is a decay of roughly \textbf{5.9 orders of magnitude in 12{,}000
steps} ($\approx$3.1B tokens), and it happens inside phase 2's first
block---the one whose mixture is 90\% English/Spanish/wiki prose and 5\%
code. We densified the grid specifically to rule out a singleton: 38K,
40K, 44K and 48K trace a smooth monotone decay rather than an isolated
spike beside a floor. The decay is also measured entirely within one probe
regime---the whitespace skip fires on 6/6 examples at all five of these
checkpoints---so it is not an artifact of the two series trading places.

Generation dies first and faster than the probability mass. At step 38K
the model still holds $3{\times}10^{-3}$ on the trigger token but already
emits 0/5 well-formed calls, and it stays at 0/5 for every subsequent
checkpoint of phase 2 and phase 3. The ability to \emph{produce} the
structure is gone roughly 10{,}000 steps before the mass finishes draining
away---which is precisely why we run the strict generation check alongside
the probe rather than treating the probe as a proxy for it.

Two caveats are load-bearing and we state them before drawing the
inference. \emph{First, the rising side is unobservable.} Step 36K is the
earliest archived checkpoint of the entire 1B pretraining run---phase 1
has none---so we can see the prior decaying but cannot see how high it
went, when it was installed, or whether 36K is near its peak or already
well down from it. \emph{Second, the probe set is contaminated for this
comparison.} All three well-formed emissions at 36K are byte-exact copies
of their training completions and none generalizes, which means the pinned
examples from \texttt{tool\_sft\_mini\_v1.jsonl} were at or very near the
phase-1 data. We verified that the contamination is not from phase 2---no
\texttt{tool\_sft} bucket appears in any phase-2 block mixture---but it
does mean the 36K number should be read as ``a memorized phase-1 prior
being washed out,'' not as evidence of a generalizing capability that
phase 2 destroyed. What phase 2 erased was a memorized prior; the paper's
claim that the 1B never had a \emph{generalizing} default
(\S\ref{sec:results:verbatim}) is unaffected, and if anything the 0/5
generalization at 36K corroborates it.

\begin{table*}[t]
\centering
\caption{\onebee trajectory probe. Left: phase-2 pretraining, 9 of 13
sampled checkpoints, with the 36K--48K window densified to test whether
the 36K value is a singleton (the four omitted points, 76K/116K/134K/174K,
sit on the same floor). Right: all ten archived phase-3 (tool-SFT)
checkpoints, with the
project's contemporaneous B4 harness score where one was recorded for
that checkpoint. Columns as in Table~\ref{tab:traj600m}; ``Emit'' is 0/5
at every 1B checkpoint after 36K. $^\S$ marks the first checkpoint after
the \texttt{tool\_sft} weight cut (0.20$\to$0.16, applied to the mixture
between steps 8K and 10K); the weight was restored to 0.20 for the
continuation running from step 10K onward.}
\label{tab:traj1b}
\footnotesize
\begin{tabular}{@{}rccccc@{\hskip 2.5em}rccccc@{}}
\toprule
\multicolumn{6}{c@{\hskip 2.5em}}{\textbf{Phase 2} (web-heavy pretraining)} &
\multicolumn{6}{c}{\textbf{Phase 3} (dedicated tool-SFT)} \\
\cmidrule(r{2em}){1-6}\cmidrule(l){7-12}
Step & $\bar P_{\mathrm{skip}}$ & $\bar P_{\mathrm{lit}}$ & no-call & Emit & Gen. &
Step & $\bar P_{\mathrm{skip}}$ & $\bar P_{\mathrm{lit}}$ & no-call & Emit & B4 \\
\midrule
36K  & \textbf{0.432}        & $3.2{\times}10^{-14}$ & $2.5{\times}10^{-2}$ & 3/5 & 0 &
4.0K   & $3.7{\times}10^{-3}$ & $3.7{\times}10^{-3}$ & $5.2{\times}10^{-3}$ & 0/5 & --- \\
38K  & $3.0{\times}10^{-3}$  & $1.0{\times}10^{-9}$  & $1.3{\times}10^{-4}$ & 0/5 & --- &
5.7K   & $1.0{\times}10^{-2}$ & $1.0{\times}10^{-2}$ & $8.5{\times}10^{-3}$ & 0/5 & 0.020 \\
40K  & $4.0{\times}10^{-4}$  & $3.8{\times}10^{-6}$  & $4.0{\times}10^{-5}$ & 0/5 & --- &
6.0K   & $\mathbf{1.4{\times}10^{-2}}$ & $1.4{\times}10^{-2}$ & $9.1{\times}10^{-3}$ & 0/5 & --- \\
44K  & $6.2{\times}10^{-7}$  & $2.1{\times}10^{-6}$  & $2.4{\times}10^{-7}$ & 0/5 & --- &
8.0K   & $1.2{\times}10^{-2}$ & $1.2{\times}10^{-2}$ & $1.4{\times}10^{-2}$ & 0/5 & 0.075 \\
48K  & $5.9{\times}10^{-7}$  & $3.3{\times}10^{-5}$  & $1.9{\times}10^{-7}$ & 0/5 & --- &
10.0K$^\S$ & $4.8{\times}10^{-4}$ & $4.8{\times}10^{-4}$ & $2.5{\times}10^{-8}$ & 0/5 & 0.020 \\
58K  & $1.2{\times}10^{-5}$  & $1.9{\times}10^{-5}$  & $1.4{\times}10^{-7}$ & 0/5 & --- &
12.0K  & $4.2{\times}10^{-4}$ & $4.2{\times}10^{-4}$ & $1.7{\times}10^{-7}$ & 0/5 & --- \\
96K  & $4.3{\times}10^{-8}$  & $4.3{\times}10^{-8}$  & $7.7{\times}10^{-9}$ & 0/5 & --- &
13.0K  & $4.3{\times}10^{-4}$ & $4.3{\times}10^{-4}$ & $2.2{\times}10^{-8}$ & 0/5 & 0.030 \\
154K & $6.1{\times}10^{-7}$  & $3.8{\times}10^{-6}$  & $5.2{\times}10^{-7}$ & 0/5 & --- &
14.0K  & $5.7{\times}10^{-5}$ & $5.8{\times}10^{-5}$ & $2.1{\times}10^{-8}$ & 0/5 & \textbf{0.085} \\
192K & $8.4{\times}10^{-7}$  & $8.1{\times}10^{-7}$  & $3.3{\times}10^{-10}$ & 0/5 & --- &
16.0K  & $2.6{\times}10^{-4}$ & $2.7{\times}10^{-4}$ & $1.4{\times}10^{-8}$ & 0/5 & 0.010 \\
     &                       &                       &                      &     &     &
20.0K  & $8.8{\times}10^{-5}$ & $8.8{\times}10^{-5}$ & $6.3{\times}10^{-9}$ & 0/5 & --- \\
\bottomrule
\end{tabular}
\\[2pt]
{\footnotesize The B4 column is drawn from the project's phase-3
checkpoint-tracking record, a different harness invocation from the
$0.650$ reproduced for this lineage in Table~\ref{tab:b15}; that the
same lineage carries two recorded B4 values differing by roughly
$8\times$ is itself consistent with \S\ref{sec:results:harness}'s verdict
on that metric, and we use the tracking series here only for
\emph{within-series} comparison, never as an absolute score.}
\end{table*}

\paragraph{The 1B, phase 3: a cliff with a known cause, and no recovery.}
Phase 3 (right panel) is not monotone. $\bar P_{\mathrm{skip}}$ rises from
$3.7{\times}10^{-3}$ at step 4K to a peak of $1.4{\times}10^{-2}$ at 6K,
holds at 8K, and then falls $24\times$ to $4.8{\times}10^{-4}$ by step
10K. That cliff has a documented cause rather than requiring one:
the project's own checkpoint record shows the \texttt{tool\_sft} bucket
being cut from 0.20 to 0.16 of the SFT mixture (and \texttt{reasoning}
from 0.15 to 0.11) to make room for new buckets in exactly that window,
and records B4 falling over the same three checkpoints, 0.075 $\to$ 0.045
$\to$ 0.020. Probe and harness agree here, which is worth saying plainly:
the instruments are not always in conflict, and when a curriculum change
is large enough both register it.

The mixture was restored---\texttt{tool\_sft} back to 0.20 for the
continuation from step 10K---and the harness partially recovered. The
structured prior did not. It ends phase 3 at $8.8{\times}10^{-5}$ at step
20K, the final phase-3 checkpoint, two orders of magnitude below its own
step-8K value and consistent with the $\sim$$10^{-4}$--$10^{-5}$ endpoint
already reported in Table~\ref{tab:verbatim}. Generation never returns at
all: 0/5 well-formed calls at all ten phase-3 checkpoints.

Nor is the early phase-3 rise a partial rescue. In the 4K--8K window the
mean-to-no-call ratio is 0.70--1.5, and at both 4K and 8K the
\emph{suppression} example carries higher probability than the call
examples---the same unconditional base-rate signature as the 600M, at a
sixtieth of the magnitude. The peak of the entire phase-3 curve
($1.4{\times}10^{-2}$) sits $61\times$ below the \emph{worst} point of the
600M curve (0.858 at step 1K). Interestingly, after step 10K the 1B does
acquire context sensitivity---the ratio jumps to $10^{3}$--$10^{4}$ as the
no-call probability collapses faster than the call probability---but it
acquires it at an absolute level where no emission ever occurs. Phase 3
taught this model when \emph{not} to call a tool long before it taught it
how to call one.

\paragraph{A direct dissociation between the harness and the probe.}
Six phase-3 checkpoints carry both a probe value and a recorded B4 score,
and at the two extremes they move in opposite directions. Step 14K holds
B4 $=$ 0.085, the highest value recorded anywhere in the 1B's history,
while the probe sits at $5.7{\times}10^{-5}$, its \emph{lowest} value in
all of phase 3---$201\times$ below step 8K. Two thousand steps later B4
collapses to 0.010, its historic floor, while the probe rebounds
$4.6\times$. The project's contemporaneous reading of the 14K peak was
``dilution of exposure, not catastrophic forgetting''; the probe says that
at the moment the lenient harness recorded its best-ever tool-use number,
the model's prior toward emitting a tool call was at its weakest, and it
emitted 0/5 calls under maximally favorable conditions. We decline to fit
a correlation to six points, and we note that the project's own record
flags both the 14K peak and the 16K floor as single-seed volatility not to
be over-interpreted individually. That caution cuts one way here: the
probe is greedy and deterministic, so the volatility lives on the harness
side of the dissociation. The conservative statement is the one the paper
already makes---B4 for this model is not measuring what its name
says---now supported by a within-model, within-run instance rather than
only the cross-model one of \S\ref{sec:results:harness}.

\paragraph{Verdict on the pre-registered prediction.}
The prediction was that ``the 1B's structured perplexity stalls or worsens
through its web-heavy phase 2 and is not rescued by phase 3.'' Both halves
hold, and the first holds more strongly than stated: phase 2 did not stall
a prior that was never there, it erased one that was, by nearly six orders
of magnitude, inside its most prose-dominated block. Phase 3 rescued
nothing---its apparent early rise is unconditional base rate rather than
capability, its one clear curriculum-driven movement is downward, and the
prior finishes two orders of magnitude below its own phase-3 peak even at
the checkpoint where the lenient harness reports the best tool-use score
in the model's recorded history. The 600M half of the prediction
(``$P(\texttt{<|tool\_call|>})$ rises as in-mix SFT accumulates'') is
confirmed only in its endpoint: the rise had already happened before the
first archived checkpoint, so what we observe is saturation, not
emergence.

\subsection{Generalization under novel entities and register}
\label{sec:results:generalization}

The paper's two starkest numbers have so far rested on the smallest
samples: verbatim reproduction on $n{=}6$ and the novel-prompt battery on
$n{=}8$. This subsection reports the scaled-up replacement for both
(pre-registered as F2 in \S\ref{sec:discussion:followups}), run on three
checkpoints: the 600M at step 154K, the pre-remediation 1B, and the
repaired 1B of \S\ref{sec:results:repair}. The last two are the reproduced
repair's parent and child, so this subsection's pre- and post-repair
columns are a matched pair within one lineage---which is what licenses the
paired test below. They are not the same pre-remediation checkpoints as
Table~\ref{tab:verbatim}'s phase-3 and phase-4b rows; all of these
checkpoints exhibit the same total absence of the format, but the
substitution is declared rather than glossed over (L12).

\paragraph{Verbatim reproduction at full corpus scale.}
Replaying the Level-3 diagnostic over all 269 deduplicated, decontaminated
corpus rows reproduces the qualitative endpoint result of
Table~\ref{tab:verbatim} and sharpens it. The 600M emits a well-formed call
on $0.926$ of rows $[0.892, 0.955]$; the pre-remediation 1B on $0.100$
$[0.067, 0.138]$; the repaired 1B on $0.959$ $[0.933, 0.981]$. The
$n{=}6$ and $n{=}4$ endpoints were not flukes of example selection.

What scale adds is a split the small sample could not show. Among
well-formed emissions, the 600M reproduces the training completion exactly
on only $0.235$ of rows---it composes a different, contextually sensible
argument on \textbf{182 of 238}---whereas the repaired 1B reproduces
exactly on $0.988$, generalizing on \textbf{3 of 246}. On prompts drawn
from its own fine-tuning corpus the repaired 1B is very largely reciting.
That is precisely the ambiguity \S\ref{sec:results:repair} flagged and
declined to resolve on $n{=}4$, and at $n{=}269$ it resolves against the
repaired model. (We do not report an exact-copy rate for the
pre-remediation 1B: its denominator is the 5 rows on which it produced a
parseable call at all, of which 3 were exact. The rate is uninformative and
we give the counts instead.)

\paragraph{A generalization battery over unseen entities.}
Whether that recitation is the whole story is what the battery decides. We
built 238 prompts over entity pools with no lexical overlap with the
corpus---reserved-range IPs, \texttt{CVE-2025-*} identifiers, randomly
generated hashes, invented domains---crossed with paraphrase templates by
true Cartesian sampling without replacement. 166 prompts expect a call,
distributed roughly flat across the five tools rather than following the
corpus's 66\% \texttt{bash\_exec} skew; 72 expect suppression. Eight
prompts reconstructed from \S\ref{sec:results:repair}'s original battery
are retained as a textual anchor and marked as such.

Scoring is decomposed rather than binary, because a single pass/fail
conflates failures we need to separate. We record, per item: whether a call
was opened at all (\emph{decision}); whether it parsed with
\texttt{name} and \texttt{args} (\emph{well-formed}); whether the tool was
the right one (\emph{tool}); whether the required argument keys were
present and non-empty (\emph{shape}); and---for the four fixed-schema
tools---whether those keys carried the \emph{expected novel value}
(\emph{entity}). A headline pass requires the tool, the shape, and the
entity; for \texttt{bash\_exec}, whose argument is free-form, it requires
the command to match the expected command family instead.

\begin{table}[t]
\centering
\caption{Generalization battery, 166 call-expecting prompts over entities
absent from the training corpus. Axes are nested: each is conditioned on
the previous one being available, so \emph{entity} is scored only on the
134--135 fixed-schema items and \emph{soft miss} only on the 30--31
\texttt{bash\_exec}-family items carrying a command regex. Brackets are
95\% Wilson intervals.}
\label{tab:genaxes}
\footnotesize
\begin{tabular}{@{}lccc@{}}
\toprule
Axis & \model & 1B pre-rep. & 1B repaired \\
\midrule
Decision      & 1.000 & 0.006 & 1.000 \\
Well-formed   & 0.994 & 0.006 & 1.000 \\
Tool correct  & 0.776 & ---   & 0.717 \\
Arg.\ shape   & 0.855 & ---   & 0.819 \\
Entity match  & 0.560 & ---   & \textbf{0.667} \\
\midrule
\textbf{Headline pass} & 0.428 & 0.000 & \textbf{0.536} \\
 & \footnotesize[.355,.504] & \footnotesize[.000,.023] & \footnotesize[.460,.610] \\
\midrule
Soft miss     & 0.767 & n/a   & \textbf{0.484} \\
\bottomrule
\end{tabular}
\end{table}

\paragraph{The repaired 1B is ahead, and the axes say why.}
Table~\ref{tab:genaxes} gives the result. The pre-remediation 1B opens a
call on 1 of 166 prompts and passes none: the battery reproduces its total
absence of the format, as expected. Between the other two, the repaired 1B
passes $0.536$ $[0.460, 0.610]$ against the 600M's $0.428$ $[0.355,
0.504]$. Because both models answer the same 166 items, the comparison is
paired: 27 items pass for the repaired 1B alone against 9 for the 600M
alone (exact McNemar $p = 0.0039$).

The axis decomposition locates the difference, and it is not where the
verbatim result would predict. The 600M is \emph{better} at choosing among
the five tools (0.776 vs.\ 0.717) and at producing a structurally complete
argument object (0.855 vs.\ 0.819). It loses on the one axis that tests
what the battery was built for: extracting an entity it has never seen
(0.560 vs.\ 0.667). So the two headline findings point in opposite
directions---on corpus prompts the repaired 1B recites and the 600M
composes; on novel entities the repaired 1B copies the unseen string out of
the prompt more reliably than the 600M does. Recitation on in-distribution
text and faithful slot-filling on out-of-distribution text are not the same
capability, and this pair of experiments separates them.

The \emph{soft miss} row is where the repaired model wins by the widest
margin, on only 30--31 items (L11d), and in the direction the repair narrative predicts. Among
items where both the tool and the argument shape are right but the specific
shell command is wrong, the 600M errs on $0.767$ of them against the
repaired 1B's $0.484$: when the repaired 1B commits to
\texttt{bash\_exec}, it picks a defensible command about twice as often.

\begin{table}[t]
\centering
\caption{Headline pass by template family. \texttt{terse\_control}
re-asks the four fixed-schema tools in the corpus's own clipped register
(``\texttt{detalle CVE-2025-3187}''); \texttt{anchor} is the reconstructed
$n{=}8$ battery of \S\ref{sec:results:repair}. The pre-remediation 1B
scores 0.000 in every family and is omitted.}
\label{tab:genfamily}
\footnotesize
\begin{tabular}{@{}lrcc@{}}
\toprule
Family & $n$ & \model & 1B repaired \\
\midrule
\texttt{nvd\_get\_cve}   & 28 & 0.893 & \textbf{1.000} \\
\texttt{anchor}          &  7 & 0.714 & 0.857 \\
\texttt{cisa\_kev\_check}& 28 & \textbf{0.607} & 0.571 \\
\texttt{otx\_check\_ioc} & 27 & 0.444 & 0.519 \\
\texttt{bash\_exec}      & 28 & 0.214 & \textbf{0.500} \\
\texttt{terse\_control}  & 20 & 0.200 & \textbf{0.500} \\
\texttt{nvd\_search}     & 28 & 0.071 & 0.036 \\
\bottomrule
\end{tabular}
\end{table}

\paragraph{No family ordering survives across both models.}
Table~\ref{tab:genfamily} breaks the headline down by template family, and
the ranking is not stable. \texttt{nvd\_get\_cve} is near ceiling for both;
\texttt{nvd\_search} is near floor for both; in between, the 600M leads on
\texttt{cisa\_kev\_check} while the repaired 1B leads by more than
$2\times$ on \texttt{bash\_exec} and \texttt{terse\_control}. With
$n \approx 28$ per cell these individual gaps carry wide intervals and we
do not interpret any one of them; the reportable observation is the
negative one, that the aggregate near-parity of Table~\ref{tab:genaxes}
does not decompose into a consistent per-tool advantage for either model.
The \texttt{anchor} rows are reassuring in a narrower way: the repaired 1B
scores 6/7 on the reconstructed call prompts, consistent with the 7/7
correct tool choices of \S\ref{sec:results:repair}, so the scaled battery
and the original one agree where they overlap.

\texttt{nvd\_search} deserves its own sentence because its failure is
specific and shared. Both models select the tool and emit a
\texttt{query} key; what they fail to do is carry across the full
multi-word search term. Asked about \texttt{gitlab cve} they emit
\texttt{gitlab}; asked about \texttt{confluence rce} they emit
\texttt{confluence}. Under exact matching these score as failures. A
relaxed criterion crediting a non-empty substring of the expected term
raises \texttt{nvd\_search} to $0.321$ (600M) and $0.357$ (repaired 1B),
and the overall headline to $0.470$ and $0.590$---which does not change the
ordering or the paired conclusion, but does mean the absolute headline
figures should be read as the strict end of a range whose lenient end is
about four to five points higher. Head-noun extraction works; full free-text
span extraction does not.

\paragraph{Register, not language, is the surface variable that bites.}
The \texttt{terse\_control} subset re-asks the same four fixed-schema tools
in the corpus's own clipped register, holding the tool set fixed so that
register is the only thing that varies. Against the same four tools at
standard conversational register, the 600M falls from $0.505$
$[0.413, 0.596]$ to $0.200$ $[0.081, 0.416]$, while the repaired 1B is flat
($0.532 \to 0.500$). The 600M's tool-calling is the more register-brittle
of the two---a result worth stating because it is the opposite of what the
verbatim comparison would suggest, and because the terse register is the
one its \emph{own} corpus is written in.

We deliberately do \emph{not} report the battery's English-vs-Spanish
contrast as a finding, although the raw split is large (600M: $0.957$ en
vs.\ $0.580$ es). It is confounded beyond repair by construction. Every
English item comes from exactly one template per family, against four
Spanish templates averaged per family; the English subset is also
composition-skewed, with 9 of its 23 call items falling in
\texttt{otx\_check\_ioc}, whose English items the 600M passes almost
without exception (it passes 22 of the 23 English call items overall,
against 0.444 on the family as a whole in Table~\ref{tab:genfamily}).
Family-matched, the gap largely dissolves and in one family reverses
(repaired 1B, \texttt{otx\_check\_ioc}: en $0.444$ vs.\ es $0.611$). What
looks like a language effect is a template-and-composition effect;
measuring language properly would require the same templates translated
and balanced, which this battery does not have (\S\ref{sec:limitations}).

\paragraph{Suppression is the shared weakness, and near-misses break both
models.}
The 72 suppression items are where both models are worst, and they
sharpen \S\ref{sec:results:trajectory}'s single-prompt finding into a rate.
On \emph{pure} items---general security questions with no tool-shaped
content---the 600M answers without calling on $0.354$ $[0.234, 0.496]$ of
prompts, over-triggering on the remaining $0.646$; the repaired 1B reaches
$0.604$ $[0.463, 0.730]$, over-triggering on $0.333$. On \emph{near-miss}
items---prompts that name a CVE, an IP, or a shell command while asking a
purely conceptual question about it---both collapse: the 600M passes
$0.087$ $[0.024, 0.268]$ and over-triggers on $0.913$, the repaired 1B
passes $0.174$ $[0.070, 0.371]$ and over-triggers on $0.783$. Mentioning a
tool-shaped entity is close to sufficient to fire a call in both models,
whatever the sentence around it asks for.

The pre-remediation 1B's suppression column must be read as the artifact it
is. It ``passes'' $0.562$ of pure and $0.565$ of near-miss items with an
over-trigger rate of exactly $0.000$---not because it discriminates, but
because it emits a tool call on 1 of 238 prompts in the entire battery. A
model that never calls a tool scores perfectly on the no-call axis. This is
the clearest single illustration of why the suppression axis is scored
jointly with a non-degenerate-answer requirement and never on its own, and
why \S\ref{sec:results:verbatim}'s call-side diagnostics carry the weight
they do.

\subsection{Ablating the repair recipe: corpus and learning rate as a
factorial}
\label{sec:results:factorial}

\S\ref{sec:results:mech} eliminated two of recipe C's four bundled
ingredients---the warm-start point, struck by the accident of two parents,
and the embedding table, struck by direct measurement showing it was
numerically inert. This subsection runs the factorial that separates the
two that remain: corpus \emph{diversity} and \emph{learning rate}.

\paragraph{Design.} Five runs, all warm-started from the same surviving
substitute parent used throughout \S\ref{sec:results:repair}
(\texttt{phase4a\_v0\_real\_llm}), all at effective batch 32, and all
trained for the same 2{,}202 steps. Equal step counts are enforced by
setting the step budget directly rather than by epochs, so that the narrow
and diverse arms take the same number of optimizer steps despite differing
in corpus size---without this the corpus factor would be confounded with
the amount of training. The narrow corpus is the tool-SFT mini set plus a
10k cyber-reasoning sample (12{,}801 unique examples); the diverse corpus
adds an open-assistant mixture (23{,}502 unique examples) and is recipe
C's actual blend. High LR is $1{\times}10^{-5}$ (recipe C's) and low LR is
$2{\times}10^{-6}$ (recipe B's). The
diverse/high-LR cell---recipe C's own corner---is run twice under different
seeds, so that between-arm differences can be read against a within-cell
one.

\paragraph{Numerical regime: this factorial is internally comparable and
externally is not.} The five runs were executed on a Turing-class T4, which
has no bf16 tensor cores, in fp16 with fp32 master weights, 8-bit AdamW and
gradient checkpointing. The recipe-C reproduction whose numbers appear in
\S\ref{sec:results:repair} and \S\ref{sec:results:mech} ran in bf16 on an
A10G, and every repaired-1B number outside this subsection comes from
that run. All five arms here share one regime and are therefore comparable to
each other; none of them is bit-comparable to the published reproduction,
and none of them replaces its numbers. Given that \S\ref{sec:results:mech}'s
central finding was itself about an update underflowing bf16 storage, the
precision change is not a detail we can wave through: an 8-bit optimizer
state and a different mantissa change exactly the quantity that analysis
turned on. We therefore read this factorial as a self-contained experiment
about corpus and learning rate under one fixed regime, not as a further
measurement of recipe C.

\paragraph{Verbatim reproduction does not separate the arms.} On the
$n{=}269$ deduplicated verbatim protocol of
\S\ref{sec:results:generalization}, all five arms are at or within noise of
ceiling: well-formed $1.000$ for narrow/high, narrow/low and
diverse/high(s42), $0.9963$ $[0.989, 1.000]$ for diverse/low and
diverse/high(s43); exact-copy $1.000$ everywhere except diverse/low at
$0.9960$. This is the expected and uninformative result. Verbatim
reproduction is the in-distribution instrument, and at a full 2{,}202 steps
every cell converges to near-total recitation of its own training corpus
regardless of corpus or learning rate. Everything that follows is the
generalization battery.

\begin{table}[t]
\centering
\caption{F3 factorial: 238-prompt generalization battery, scored on the
same axes as Table~\ref{tab:genaxes}. Denominators follow the scorer's own
conditioning: headline over the 166 call items, tool over the
well-formed call items (165--166), entity over the
134--135 fixed-schema items, soft miss over the 31 command-regex items,
suppression over the 72 no-call items. The two right-hand columns are
complements of each other up to a single degenerate answer and are not
independent evidence. Best in column in \textbf{bold}; for soft miss and
over-triggering the better value is the \emph{lower} one, as in
Table~\ref{tab:genaxes}.}
\label{tab:f3factorial}
\footnotesize
\begin{tabular}{@{}lcccccc@{}}
\toprule
Arm & Head. & Tool & Entity & Soft & Supp. & Over \\
 & pass & corr. & match & miss & pass & trig. \\
\midrule
narrow / high LR    & \textbf{0.608} & 0.783 & \textbf{0.748} & 0.419 & 0.458 & 0.542 \\
narrow / low LR     & 0.542 & 0.685 & 0.702 & \textbf{0.387} & 0.375 & 0.625 \\
diverse / low LR    & 0.584 & 0.759 & 0.719 & 0.452 & \textbf{0.583} & \textbf{0.403} \\
diverse / high (s42)& 0.602 & \textbf{0.795} & 0.741 & 0.452 & 0.514 & 0.486 \\
diverse / high (s43)& 0.602 & 0.771 & 0.726 & \textbf{0.387} & 0.500 & 0.500 \\
\bottomrule
\end{tabular}
\end{table}

\paragraph{How these are tested.} All five arms are scored on the
\emph{same} 238 prompts, so every comparison is item-matched and is tested
as such---exact McNemar on the discordant pairs, with paired-bootstrap
intervals on the difference---rather than as two independent proportions.
This matters for power: the unpaired reading of a 166-item battery would
call almost nothing here significant. Against that, the factorial runs 36
tests---six contrasts (the corpus contrast at low LR and at high LR against
each high-LR seed; the LR contrast in the narrow corpus and in the diverse
corpus against each high-LR seed) on the six axes of
Table~\ref{tab:f3factorial}---so we report Benjamini--Hochberg control at
$q{=}0.05$ across all 36 and treat only the survivors as findings. Four
survive: the low-LR corpus contrast on suppression, over-triggering and
tool correctness (Finding~1), and the narrow-corpus LR contrast on tool
correctness (Finding~3).

\paragraph{The seed replicate says the noise floor is not where the
aggregate suggests.} The two diverse/high-LR arms differ only in seed, and
their headline pass rates are identical to four decimals ($0.6024$ both).
That identity is a coincidence of cancellation, not evidence of
determinism: the two arms disagree on six individual items, three in each
direction. On tool correctness they differ by $+0.0241$ (4 discordant
items, all one way), which is as large as the entire learning-rate effect
measured inside the diverse corpus. The honest summary is that run-to-run
noise on this battery is small in aggregate but not negligible at the scale
of the effects the factorial is trying to resolve, and that an aggregate
rate matching across seeds should not be read as a stable cell.

\paragraph{Finding 1: at low LR, the diverse arm suppresses far better than
the narrow arm---in one run per cell.} The largest contrast in the
factorial is not about headline accuracy. Holding LR at $2{\times}10^{-6}$,
moving from the narrow to the diverse corpus raises suppression pass from
$0.375$ to $0.583$ ($+0.208$, CI $[+0.125, +0.306]$, $p{=}6{\times}10^{-5}$)
and cuts over-triggering from $0.625$ to $0.403$ ($-0.222$,
CI $[-0.319, -0.125]$, $p{=}3{\times}10^{-5}$).
The discordance is entirely one-directional---15 items flip toward the
diverse arm and 0 toward the narrow one on suppression, 16 to 0 on
over-triggering---which is why an effect this size clears correction on a
72-item subset. Tool correctness moves too, $0.685 \to 0.759$
($p{=}0.004$). Suppression pass and over-triggering are the same
underlying behavior counted twice, so we report these as one finding, not
three. Both cells behind it were run once, and the partial seed replicate
reported below does not reproduce its suppression half; we therefore hold
that half as a hypothesis, not a finding (see below).

\paragraph{Finding 2: at high LR, the corpus makes no measurable
difference.} The same corpus contrast run at $1{\times}10^{-5}$ is flat on
every axis: headline $0.602$ vs.\ $0.608$, tool $0.795$ vs.\ $0.783$,
suppression $0.514$ vs.\ $0.458$, no test below $p{=}0.42$, and the
headline difference ($-0.006$) is smaller than the seed-to-seed
disagreement measured above. Whatever the open-assistant mixture
contributes at low LR, it is not detectable once the learning rate is high
enough.

\paragraph{Finding 3: the learning rate matters in the narrow corpus and is
not established in the diverse one.} Within the narrow corpus, raising LR
improves tool correctness from $0.685$ to $0.783$ ($+0.098$,
CI $[+0.049, +0.152]$, 18 discordant items to 2, $p{=}4{\times}10^{-4}$,
the third BH survivor) and headline pass from $0.542$ to $0.608$
($+0.066$, $p{=}0.007$, which clears the uncorrected threshold but not
BH). Within the diverse corpus the same contrast is inconsistent across the
two seeds---tool correctness $+0.036$ ($p{=}0.031$) against seed 42 but
$+0.012$ ($p{=}0.73$) against seed 43, headline $+0.018$ ($p{=}0.375$)
against both---so we do not claim a learning-rate effect there.

\paragraph{The two ingredients look substitutable rather than additive, with
one cell carrying the whole result.} Reading the three findings together:
either ingredient alone reaches a headline pass around $0.60$, having
neither drops it to $0.542$, and having both adds nothing over having one.
We do not report this as an interaction effect with a confidence interval:
three of the four cells were run to completion once, so any such estimate
would carry item-sampling variance and no run-level variance at all, and
the cells where run-level variance \emph{is} observable (the seed replicate
at high LR, and the partial narrow/low-LR replicate below) produced swings
comparable to the effects an interaction term would be trying to resolve. The substitutability reading above is a
description of five checkpoints, not an estimate of a population quantity,
and we leave it at that.

\paragraph{A partial seed replicate does not reproduce the suppression half
of Finding~1.} Finding~1 rests on exactly the two cells that were run once.
Seed replicates of both low-LR cells were planned. The diverse/low-LR
replicate was never started; the narrow/low-LR replicate (seed 43) was
interrupted at step 880 of 2{,}202 (40\% of training). Its training log
matches the seed-42 run in everything but the seed (same parent, corpus
files and record counts, LR, warmup, step budget, batch and precision);
since both start from the same weights, the seed acts through data order.
We evaluated its step-880 checkpoint alongside the step-880 checkpoints of
the original narrow/low and diverse/low arms, with the same CPU fp32 greedy
path and scorer as Table~\ref{tab:f3factorial}
(Table~\ref{tab:f3step880}). Verbatim reproduction again does not separate
them: well-formed and exact-copy are $1.000$ for both narrow seeds, and
$0.985$ and $0.996$ for the diverse arm.

\begin{table}[t]
\centering
\caption{Matched-step check at step 880 of 2{,}202 (40\% of training):
the two low-LR arms of Table~\ref{tab:f3factorial} and an interrupted
second seed of the narrow/low-LR cell. Same battery, scorer and
denominators as Table~\ref{tab:f3factorial}. These values are comparable
with each other, not with the final-step table.}
\label{tab:f3step880}
\footnotesize
\begin{tabular}{@{}lcccccc@{}}
\toprule
Arm @ step 880 & Head. & Tool & Entity & Soft & Supp. & Over \\
 & pass & corr. & match & miss & pass & trig. \\
\midrule
narrow / low (s42)  & 0.536 & 0.691 & 0.701 & 0.419 & 0.389 & 0.611 \\
narrow / low (s43)  & 0.548 & 0.685 & 0.687 & 0.355 & 0.569 & 0.431 \\
diverse / low (s42) & 0.578 & 0.759 & 0.711 & 0.452 & 0.556 & 0.431 \\
\bottomrule
\end{tabular}
\end{table}

On suppression, the two seeds of the \emph{same} cell differ by $+0.181$
(CI $[+0.097, +0.278]$; 13 items to 0, $p{=}2{\times}10^{-4}$)---as much as
the corpus contrast at the same step (diverse vs.\ narrow seed 42:
$+0.167$, 12 to 0, $p{=}5{\times}10^{-4}$)---and the second narrow seed is
indistinguishable from the diverse arm ($-0.014$, 3 to 4, $p{=}1.0$).
Tool correctness behaves differently: the two narrow seeds agree ($0.691$
and $0.685$; 2 items to 3), and the diverse arm exceeds both (13 to 2,
$p{=}0.007$; 13 to 1, $p{=}0.002$). Headline pass separates no pair
(smallest $p{=}0.09$). These are 18 further exact McNemar tests (three
contrasts on six axes, uncorrected $p$ above); under Benjamini--Hochberg
at $q{=}0.05$ within them, the survivors are exactly six: the
suppression and over-triggering versions of the two contrasts against
narrow seed 42, and the two tool-correctness contrasts. Two observations bear on reading step 880 against the final
step. In the two seed-42 arms, step 880 already sits close to step 2{,}202
(suppression $0.389$ vs.\ $0.375$ narrow, 1 discordant item; $0.556$
vs.\ $0.583$ diverse, 2 items), so the step-880 picture is not a transient
of those runs; whether the seed-43 run would have moved over its remaining
60\% is unmeasured. And run-level variance is not uniform across cells:
the two high-LR diverse seeds disagree on 3 suppression items (2 to 1),
the two low-LR narrow seeds on 13.

Our reading is that, at 40\% of training, seed-to-seed variation within
the narrow/low-LR cell is as large as the corpus contrast on suppression.
The suppression half of Finding~1 is therefore not replicated, and we hold
it as a hypothesis: a single seed of the narrow/low cell happened to
suppress poorly, and whether diversity shifts the \emph{distribution} of
suppression outcomes at low LR is untested. It is not refuted either---this
is one extra seed, of one of the two cells, at 40\% of training, on 72
items. The tool-correctness half survives the only check available,
against two narrow seeds, though the diverse/low-LR cell is still a
single run.

\paragraph{The narrow/low-LR cell installs the call format.} One result
of the factorial bears on the repair narrative rather than on the two
ingredients. The narrow/low-LR cell is close to recipe B of
Table~\ref{tab:recipes}: the same learning rate and a similar narrow
corpus, but 2{,}202 steps instead of 486, a different parent, and this
factorial's fp16 regime. It is the weakest cell, but it is not a failure:
it opens a call on all 166 call items, is well-formed on 0.994 of them,
reproduces its corpus at ceiling, and reaches a headline pass of $0.542$.
(For scale only, since the regimes differ: the un-fine-tuned 600M scores
$0.428$ on the same battery, Table~\ref{tab:genaxes}.) The same arm's
intermediate checkpoint at step 440---closer to recipe B's 486 steps, but
still not recipe B, which also differed in parent, precision and frozen
embeddings---already opens a call on all 166 call items and passes $0.554$
of them, while suppressing on only $0.250$ of no-call items (step 880:
$0.389$; 10 items to 0, $p{=}0.002$). We cite it as a proxy for scale, not
as a measurement of the lost run. Whatever recipe B's true outcome was, a
narrow corpus at its learning rate is not by itself enough to prevent the
repair. Together with Finding~2, this also
qualifies the prediction that opened \S\ref{sec:results:repair}: on the
call side, a narrow corpus does as well as the diverse one once the
learning rate is high; at low learning rate diversity is associated with
better tool correctness, and its apparent suppression advantage does not
survive a second narrow seed.

What the factorial does establish, and what it does not: the completed
narrow/low-LR run---the corner furthest from recipe C, and approximately
recipe B run for longer---is the weakest arm on
headline pass, tool correctness and suppression alike, which is
corroboration that recipe C's diverse/high-LR corner was the right choice
for reasons other than luck. The one axis where it is not weakest is soft
miss, where it is nominally best at $0.387$; we decline to read anything
into that, since soft miss is scored on 31 items, no contrast on it comes
close to significance anywhere in the factorial, and its two best cells are
narrow/low and diverse/high(s43), which share neither factor. The step-880
replicate adds a caution to ``weakest'': on suppression, the second seed of
the same cell was not weak at all. Beyond that, the defensible claim is
narrow: at high learning rate the corpus buys nothing measurable; at low
learning rate the diverse arm is better on tool correctness against both
narrow seeds; and the large suppression gain that first looked like the
factorial's main result is, on current evidence, within the run-to-run
variation of the narrow/low cell. If the completed replicates confirmed
it, it would be compatible with the routing account of
\S\ref{sec:results:mech}---diversity cannot substitute for an update large
enough to move mid-network weights, but it could partly compensate for one
that is not---but it would not be a test of it. We do not claim the
factorial identifies a single dominant ingredient, and we no longer claim
that the corpus governs suppression.

\subsection{Tool SFT of the 600M, and a null distillation ablation}
\label{sec:results:sft600m}

Everything above uses the 600M exactly as pretraining left it. Two
questions follow naturally: does a dedicated tool SFT give the 600M the
call/no-call gate it lacks, and does training data distilled from a much
larger model help at this scale? We ran one SFT ablation that answers the
second directly and bears on the first. It is the only experiment in this
paper that fine-tunes the 600M, and none of its numbers appear elsewhere.

\paragraph{Setup.} Both arms start from the step-154K checkpoint and share
every hyperparameter: 1{,}200 steps, learning rate
$1.5{\times}10^{-5}$, sequence length 2{,}048, effective batch 32, one
seed, one L4 GPU. The \emph{baseline} arm trains on 17{,}801 records: the
2{,}801-example tool corpus plus 15{,}000 domain-reasoning traces. The
\emph{distilled} arm adds about 4{,}443 records ($\approx$20\% of a
22{,}244-record mix) generated by an open-weight teacher (GLM-5.3), grounded
in real NVD CVE records~\cite{nvd} and in public Sigma detection rules,
and filtered by a gate on the language of the reasoning. One distillation
stream was still running at launch, so the mix contains only what existed
then. An earlier version of this ablation used a 1{,}024-token context,
which silently truncated about a fifth of the mix, nearly all of it the
new distilled examples; it is superseded and not reported. Both arms are
scored on the 238-prompt battery of
\S\ref{sec:results:generalization} with the same scorer.

\begin{table}[t]
\centering
\caption{Tool SFT of the 600M, one seed per arm, on the 238-prompt battery.
The first column is the un-fine-tuned 600M, copied from
Table~\ref{tab:genaxes} and \S\ref{sec:results:generalization}; it is
not paired-tested against the SFT arms here. Tool is conditioned on a
well-formed call and entity on the fixed-schema items, as in
Table~\ref{tab:genaxes}; paired tests use the items scored in both arms.
``Discordant'' counts items passing only in the baseline arm vs.\ only in
the distilled arm, with the exact McNemar $p$.}
\label{tab:sft600m}
\footnotesize
\begin{tabular}{@{}lcccc@{}}
\toprule
 & 600M & SFT & SFT $+$ & Discordant \\
 & (no SFT) & baseline & distilled & ($p$) \\
\midrule
Headline pass (166)    & 0.428 & 0.548 & 0.548 & 5 / 5 (1.00) \\
Well-formed            & 0.994 & 0.934 & 0.922 & 7 / 5 (0.77) \\
Tool correct           & 0.776 & 0.794 & 0.817 & 1 / 6 (0.13) \\
Entity match           & 0.560 & 0.714 & 0.712 & 2 / 4 (0.69) \\
\midrule
Suppression pass (72)  & ---   & 0.014 & 0.056 & 0 / 3 (0.25) \\
\quad pure (48)        & 0.354 & 0.000 & 0.063 & \\
\quad near-miss (23)   & 0.087 & 0.043 & 0.043 & \\
Over-trigger (72)      & ---   & 0.972 & 0.944 & 2 / 0 (0.50) \\
\bottomrule
\end{tabular}
\end{table}

\paragraph{The distilled data has no measurable effect.} The two arms
pass exactly the same fraction of call items, with five items flipping in
each direction; the paired 95\% bootstrap interval on the headline
difference is $[-0.036, +0.036]$. No axis differs significantly. The
largest movements, tool correctness (6 items to 1) and suppression (3 to
0), favor the distilled arm but are far from significance even before any
correction. This is a single seed at small $n$, so a small effect cannot
be excluded; what the data exclude is a large one.

\paragraph{Tool SFT does not give the 600M a gate; it removes what little
suppression it had.} Against the un-fine-tuned 600M, both SFT arms pass
more call items (0.548 vs.\ 0.428) and extract the unseen entity more
often (0.71 vs.\ 0.56). This comparison is descriptive: it crosses a
training intervention and is not paired-tested here. On the no-call side
the direction reverses. The un-fine-tuned 600M answered without a call on
0.354 of the pure suppression prompts; after tool SFT the baseline arm
does so on none of 48 and calls a tool on 70 of the 72 no-call prompts.
This is what the base-rate reading of \S\ref{sec:discussion:mechanism}
predicts: SFT on a corpus in which almost every assistant turn is a call
raises the rate of calling, and nothing in it teaches when not to. It
would also match the direction of Finding~1's suppression half, which a
second narrow seed did not reproduce; this is a different model, corpus
and regime in any case, and we do not treat it as evidence either way. The
suppression pass requires a non-degenerate answer, not a correct one;
several of the few passing answers here are factually wrong.

\section{Discussion}
\label{sec:discussion}

\subsection{Thesis: composition decides default vs.\ intervention}
\label{sec:discussion:thesis}

The full record now spans four outcomes: (a) a 600M model whose only
instruction-formatted exposure was $\approx$0.1\% of a single-phase
code-heavy mix emits well-formed, generalizing tool-call JSON by default;
(b) a 1.1B sibling with a web-heavy curriculum and a dedicated
$\approx$6B-token tool-SFT phase does not, at any pre-remediation
checkpoint; (c) a diagnosis-informed recipe repairs the 1B to genuine,
generalizing tool selection in 2{,}202 steps on one rented GPU; and (d) a
factorial over that recipe's corpus and learning rate finds every cell,
including the narrow low-LR corner, installing the call format, with the
cells differing mainly in how well they suppress---an axis on which two
seeds of one cell can differ as much as two corpora. (A fifth record, a narrow
low-LR retune once reported as failing, was never validly measured;
\S\ref{sec:results:repair}.)

An earlier draft of this work, written before (c), framed the contrast as
binary: code-heavy bootstraps \emph{install} the structured-output prior and
web-heavy bootstraps do not, full stop. Outcome (c) falsifies the strong
form, and we say so plainly. The defensible form is about \emph{defaults
and difficulty}, not possibility: bootstrap composition determines whether
structured-output capability arrives for free---installed by pretraining
and activated by a trace of in-mix signal---or must be engineered in
afterward by a correctly designed intervention. And ``correctly designed''
is doing real work in that sentence, because the record shows SFT
\emph{token volume} predicted nothing: the failed dedicated phase outspent
the successful fix by roughly three orders of magnitude. What separated
recipe C from recipe A was not more exposure to tool traces but a short,
dedicated run whose data was mostly assistant turns, rather than a
minority share of a broad mixture. The factorial (outcome (d)) shows that
less of recipe C was necessary than we first assumed: context diversity
(conversational $+$ reasoning $+$ tool data) is not needed for the call
side at the high learning rate. At the low one the diverse arm suppressed
far better than the narrow arm, but a partial second seed of the narrow
cell suppressed as well as the diverse arm, so that contrast is within
run-to-run variation and remains a hypothesis.

\subsection{Why would code-heavy pretraining install the default?}
\label{sec:discussion:mechanism}

We can be more mechanistic than ``code helps,'' at the token level, and the
mechanism probe sharpens where the story must live.

\paragraph{(1) The format tokens are high-frequency in code and rare in prose.}
A tool call is, tokenwise, a sequence dominated by \texttt{\{}, \texttt{"},
\texttt{:}, \texttt{,}, \texttt{\}} and short quoted identifiers. In source
code---especially the JSON, YAML, dicts, keyword arguments, and API
literals that saturate a Stack-derived corpus~\cite{kocetkov2022stack}---these
tokens and their local $n$-gram contexts (\texttt{\{"}, \texttt{":},
\texttt{", "}) are among the most frequent items in the distribution. In web
prose they are rare and largely confined to atypical documents. A model
trained at 65\% code assigns these continuations high base probability in
neutral contexts; a model trained mostly on prose must overcome a large
log-probability deficit at every position of the emission.

\paragraph{(2) Code teaches the constraint, not just the tokens.}
Well-formedness is not a bag-of-tokens property: brackets must balance,
quotes must pair, keys must precede values. Code corpora supply billions of
examples where these constraints hold essentially without exception. This is
the same reason code-trained models do better at structured commonsense
generation when the target is expressed as code~\cite{madaan2022codelm} and
why code fractions help structured tasks broadly~\cite{aryabumi2024code}.
The 600M's generalization behavior---new, valid commands inside an
always-valid schema---is what a constraint-level prior predicts.

\paragraph{(3) The deficit is routing, not representation---but the 600M's
prior is a base rate, not routing.}
The embedding-drift result (\S\ref{sec:results:mech}) relocates where the
compositional advantage must live. The 1B's \texttt{<|tool\_call|>}
embedding was never broken, and the repair never moved it---at this
learning rate it could not have, the per-step update being smaller than the
bf16 precision the table is stored in. Because the model ties input and
output embeddings, that unmoved row is also the unembedding direction the
final hidden state is scored against: the target of routing sat fixed while
the repair took hold. What the repair built---and what recipe A failed to
build---is the mid-network \emph{routing} that steers hidden
states toward that fixed direction when the context calls for it. The
novel-prompt battery is what licenses the word ``routing'' there: the
repaired 1B decides call vs.\ no-call correctly on 8/8 prompts and selects
among five tools. At the scale of
\S\ref{sec:results:generalization} that license narrows but survives---the
repaired 1B selects the right tool on $0.717$ of 166 unseen-entity prompts
and extracts the unseen argument on $0.667$ of the fixed-schema ones, which
is routing on the call side; what does not survive is the no-call side,
discussed below.

An earlier version of this paragraph extended the same word to the 600M,
reading its default as pretraining-installed routing. The trajectory probe
(\S\ref{sec:results:trajectory}) does not support that extension and we
withdraw it. The 600M assigns the suppression example the same probability
as the call examples---mean-to-no-call ratio 1.00 across all 15
checkpoints---and emits a tool call on it at every one of them. What
pretraining installed is a high \emph{unconditional base rate} for opening
a bracketed key--value object in an assistant turn, not a
context-conditional policy that selects when to open one. This is the same
claim as (1) above, and only that claim: a model that spent 65\% of its
gradient inside structured text finds ``the continuation here is a
bracketed key--value object'' cheap in \emph{every} context, and the
in-mix SFT trace connects that cheap continuation to the chat template
without ever teaching a gate. The 600M's advantage over the 1B is
therefore real but narrower than routing: it produces well-formed,
generalizing calls by default, and it over-produces them.

This correction's weakest link used to be that our suppression evidence
was one pinned example ($n{=}1$ per checkpoint), with no power to detect a
partial or noisy gate. F2 (\S\ref{sec:results:generalization}) has since
scaled that condition to 72 held-out no-call prompts, and it confirms the
correction on the 600M: the model answers without calling on only $0.354$
of topically unrelated prompts, and on $0.087$ of prompts that merely
\emph{mention} a CVE or a shell command while asking a conceptual question.
The base-rate reading survives contact with a properly powered
suppression set. A dedicated tool SFT of the 600M does not add a gate
either (\S\ref{sec:results:sft600m}): it raises the call-side pass rate
and drives suppression on the pure prompts from 0.354 to zero.

\paragraph{Two instruments, two questions: reconciling F1 and F2 on the
repaired 1B.}
The same scaling produces an apparent tension that is worth stating
explicitly, because the two sections can be read as contradicting each
other. \S\ref{sec:results:trajectory} closes by observing that after
phase-3 step 10K the 1B's mean-to-no-call ratio jumps to
$10^{3}$--$10^{4}$, and concludes that phase 3 taught the model when
\emph{not} to call a tool long before it taught it how to call one.
\S\ref{sec:results:generalization} then reports that the repaired 1B
over-triggers on $0.333$ of pure and $0.783$ of near-miss suppression
prompts. Discrimination that looked near-perfect in one experiment looks
poor in the other.

Both are correct, because they measure different quantities. F1's ratio is
\emph{relative} and \emph{in-distribution}: it compares the probability
assigned to one pinned corpus no-call example against five pinned corpus
call examples, at a point on the curve where all six probabilities are near
zero in absolute terms and no emission occurs at all. A ratio of $10^{4}$
between two vanishing quantities describes the \emph{shape} of the
distribution over contexts the model was trained on; it says nothing about
what the model does when it can actually emit. F2's rate is
\emph{absolute} and \emph{out-of-distribution}: it asks how often a model
that now reliably emits calls withholds one on a prompt it has never seen,
about entities it has never seen. The first is a statement about context
architecture---the machinery to condition on context exists and is
correctly oriented. The second is a statement about robustness---that
machinery does not transfer to novel surface forms. A model can have a
correctly-shaped but non-transferring gate, and the repaired 1B is one.

The practical reading is that the repair installed the gate's
\emph{structure} without installing its \emph{coverage}. Note that the
repaired 1B is nonetheless the better-gated of the two models on every
suppression cell we measured (pure $0.604$ vs.\ $0.354$; near-miss $0.174$
vs.\ $0.087$), which is consistent with F1's account of where each model's
discrimination came from: the 1B has a gate that generalizes badly, the
600M has substantially no gate at all.

The routing account of the \emph{1B's} repair is untouched by any of this,
but an earlier version of this paragraph leaned on two things the record no
longer supports. It used recipe B's ``failure'' as evidence that narrow
re-exposure fits sequences without building routing; that failure was
never validly measured (\S\ref{sec:results:repair}). And it argued that
context-\emph{diverse} SFT is what builds routing; the factorial finds the
narrow corpus doing as well on the call side once the learning rate is
high, and the one suppression advantage of diversity it showed, at low
learning rate, is not reproduced by a partial seed replicate
(\S\ref{sec:results:factorial}). What survives is narrower: base rate
and gate behave as separable properties. Pretraining composition bought
the 600M the first cheaply; tool SFT, narrow or diverse, bought the 1B the
first and only part of the second, and bought the 600M none of it.

\paragraph{(4) SFT binds or builds, at very different prices.}
The install-vs-bind asymmetry of the earlier framing survives in cost form.
For the 600M, SFT had only to \emph{bind} an existing base rate for
bracketed key--value output to the trigger token---achievable with
$\sim$$10^7$ interleaved tokens. For the 1B, SFT had
to \emph{build} routing that pretraining never laid down---which failed at
$\approx$6B tokens of the wrong shape (narrow re-exposure inside a broad
mix) and succeeded at $\sim$$10^7$ tokens of the right shape (diverse
assistant-mode contexts, higher LR). The price difference between the two
models is thus not measured in tokens but in \emph{design information}: the
600M needed no diagnosis; the 1B's fix was cheap only once the failure was
correctly localized. Teams without the diagnostic would rationally have
concluded, as we nearly did, that the capability was out of reach.

\paragraph{(5) Register, generalized.}
The Nano paper's finding was that the bootstrap corpus's conversational
register dominated downstream chat behavior regardless of later
tuning~\cite{santillana2026vectrayx}. The present result is the same law one
level down, with its edges now measured: \emph{syntactic} register
(bracketed key--value text vs.\ flowing prose) dominates the
\emph{default} structured-output behavior, and later tuning overrides the
default only when it is designed against the actual failure mode. The
practical rule for small models is unchanged in spirit and sharpened in
letter: decide what the model must emit natively and make the bootstrap mix
look like it---or budget for the diagnosis-and-repair loop this paper
documents.

\subsection{Practical implications}
\label{sec:discussion:practical}
For teams building sub-1B tool-using models on small budgets:
(i) allocate code/structured-text fraction as a first-class design decision
aimed at the output format, not only at coding benchmarks;
(ii) log $P(\texttt{<|tool\_call|>})$-style first-token probes as
training-time telemetry---near-free, and \S\ref{sec:results:trajectory}
shows retrospectively that they would have flagged the 1B at phase-2 step
38K, roughly 40B pretraining tokens and an entire $\approx$6B-token
tool-SFT phase before the failure was actually noticed; log both the
literal and whitespace-skipped variants, since they disagreed by thirteen
orders of magnitude at one checkpoint in our sweep;
(iii) gate any tool-use benchmark number behind a strict structural check
(\S\ref{sec:eval:verbatim}), because keyword harnesses fail open;
(iv) when tool-SFT fails, do not conclude from volume that the capability
is unreachable---recipes A and C show failure at 6B tokens coexisting with
success at $\sim$$10^7$, and in the factorial every combination of corpus
and learning rate we tried installed the format, differing mainly in
suppression, where run-to-run variation is also large;
(v) pair every fine-tune-on-corpus-X evaluation with novel-prompt
generalization tests, since verbatim success alone is indistinguishable
from memorization.

\paragraph{Point (iii), confirmed against our own pipeline while writing
this paper.} While closing the pending B1/B4 numbers for the repaired
checkpoint (L5), the formal harness reported \texttt{tool\_call} emission
at 0/8, contradicting the strict diagnostic's 8/8 on the same checkpoint
minutes earlier. The cause was exactly the failure mode this paper is
about, one layer removed: the harness decoded generated ids with a
tokenizer library whose default silently drops tokens flagged
\texttt{special} before the string match runs, so a perfectly-formed
\texttt{<|tool\_call|>\{...\}<|/tool\_call|>} response scored as absent
tool use for a reason having nothing to do with the model. A one-line fix
(decoding without dropping special tokens) recovered 8/8. We audited the
blast radius rather than assume it: the same decode pattern, and the
mirror-image bug in a GGUF exporter that marks \texttt{<|tool\_call|>} as
a \texttt{CONTROL} token, together explain a benchmark-shaped
``0/3 literal emission'' finding recorded repeatedly during this series'
1B development and used to justify a corpus reweighting---while the
published \model results and the Nano-series numbers we checked were
produced by scoring paths that never depended on the token surviving
decode, and are unaffected. We report this as a second, independent
instance of the paper's own thesis rather than omit it: a keyword harness
failed open inside the project that named the failure mode, three layers
deep (eval script, GGUF export, and a production agent loop that shares
the pattern), and the fix in every case was the same one-line class of
error, not a modeling problem.

\subsection{Proposed follow-up experiments}
\label{sec:discussion:followups}

Ordered by cost; F1 and F2 are CPU-only, F3 needed a single small GPU,
and F4 needs real pretraining compute.

\paragraph{F1: Trajectory probes over archived checkpoints---\emph{run};
see \S\ref{sec:results:trajectory}.}
This was pre-registered here as a near-zero-cost follow-up, with the
prediction that the 600M's $P(\texttt{<|tool\_call|>})$ would rise as
in-mix SFT accumulated while the 1B's prior stalled or worsened through
web-heavy phase 2 and was not rescued by phase 3. We then ran it over 15
600M checkpoints and 23 1B checkpoints; \S\ref{sec:results:trajectory}
reports the outcome, which confirmed the 1B half more strongly than
predicted (phase 2 erased an inherited prior rather than merely failing to
build one), showed the 600M already saturated at its first archived
checkpoint, and produced two findings the prediction did not anticipate:
the 600M's default is an unconditional base rate rather than a gate
(\S\ref{sec:discussion:mechanism}), and B4 and the probe move in opposite
directions across the 1B's phase-3 checkpoints. The one component of F1
still unrun is the structured-syntax (JSON/bracket) perplexity battery,
which would test whether the base-rate account holds on ordinary
structured text rather than only at the trigger token; it costs the same
as what we ran. The remaining follow-ups below are as originally stated.

\paragraph{F2: Scale the strict diagnostics---\emph{run};
see \S\ref{sec:results:generalization}.}
This was pre-registered as extending verbatim reproduction from $n{=}6$ to
the full corpus and the generalization battery from $n{=}8$ to hundreds of
held-out prompts, with intervals, over all three checkpoints; its named
first target was the suppression condition, then a single pinned prompt
carrying the paper's base-rate claim. We ran it at $n{=}269$ (verbatim)
and $n{=}238$ (battery, of which 72 suppress).
\S\ref{sec:results:generalization} reports the outcome. It confirmed the
base-rate claim for the 600M, and produced three findings the
pre-registration did not anticipate: the repaired 1B \emph{beats} the 600M
on unseen-entity prompts (paired $p = 0.0039$) despite reciting on
in-corpus ones, so exact reproduction and novel-slot extraction dissociate;
the 600M is the more register-brittle model, losing 30 points when the same
tools are asked for in its own corpus's clipped style; and suppression
fails for both models on near-miss prompts, at $0.087$ and $0.174$.

Two components remain unrun. The battery covers neither \emph{unseen
tools} nor \emph{multi-call sequences}---every item targets one of the five
trained tools with a single call---so nothing here speaks to schema
transfer or composition. And the preamble was held fixed, so the
pre-registered sweep over how strongly the system prompt forbids a call is
still open; given that near-miss suppression is now the weakest measured
behavior in the paper, that sweep is the natural next step.

\paragraph{F3: Ablate the repair recipe---one arm resolved by measurement,
the rest run as a factorial (\emph{done};
\S\ref{sec:results:factorial}).}
Recipe C bundled four changes. Two are no longer open. The
\emph{warm-start point} was struck by accident: the recipe was run twice
from two different parent checkpoints and succeeded to the same strict
standard both times (\S\ref{sec:results:repair}). The
\emph{frozen-embedding} arm, pre-registered here in an earlier draft as the
direct test of the routing account, was resolved by measurement rather than
by a run: recipe C's embeddings were numerically inert, so the recipe
already executed with embeddings effectively frozen and the predicted
outcome is the one observed (\S\ref{sec:results:mech}). We report that as a
confirmation obtained by measuring an ingredient that never acted, not as
an ablation, and we do not claim the contrast it would have drawn.

What remains open is the ingredient pair that the elimination leaves
standing: corpus \emph{diversity} and \emph{learning rate}. An earlier
draft proposed testing these as two loose runs---narrow corpus at high LR,
diverse corpus at low LR. That design cannot separate an interaction from a
main effect, which matters here because the diagnosis predicts one
specifically (diversity should buy little at an LR too low to move
mid-network weights). We have therefore replaced it with a
\textbf{2$\times$2 factorial}---corpus $\in$ \{narrow, diverse\} $\times$
LR $\in$ \{$2{\times}10^{-6}$, $1{\times}10^{-5}$\}, all four arms
warm-started from the same parent, run for the same 2{,}202 steps---plus a
\textbf{seed replicate} of the diverse/high-LR cell, so that a
between-arm difference can be read against a within-cell one. Recipes B and
C are approximately the narrow/low and diverse/high corners already, but
neither shares a parent or a step count with the other, so we re-run all
four rather than import them.

The measurement plan is what converts this from a demonstration into an
ablation. Each of the five runs gets the \emph{full} F2 suite---verbatim
reproduction at $n{=}269$ and the 238-prompt generalization battery,
scored on the same nested axes as Table~\ref{tab:genaxes}---rather than the
$n{=}4$/$n{=}8$ instruments the original repair was judged on. The point of
the expense is inferential: every ingredient claim in this paper is
currently correlational, resting on comparisons \emph{between} models with
different lineages, corpora, and scales. Five runs differing in exactly two
declared factors, within one lineage, measured on one instrument, make the
claim causal for the first time.

All five runs have now been executed and are reported in
\S\ref{sec:results:factorial}. Two qualifications on the plan as stated
above survived contact with the run. The first is the one anticipated
parenthetically here in an earlier draft: the factorial went to a T4, whose
lack of bf16 forced fp16 with an 8-bit optimizer, so the five arms are
internally comparable but are \emph{not} bit-comparable to the bf16 recipe-C
reproduction in \S\ref{sec:results:repair}---and since
\S\ref{sec:results:mech}'s finding was precisely about an update
underflowing bf16, that substitution is disclosed rather than glossed. The
second is inferential. The design promised to make the ingredient claims
causal; what it delivers is weaker, because three of the four cells were
run to completion once and both cells with a second seed show run-to-run
movement comparable to the effects being separated. In the one case where
that was checked directly it decided the reading: the low-LR corpus
contrast on suppression, the factorial's largest effect, is matched by the
difference between two seeds of the narrow cell at step 880, so it is
reported as an unreplicated observation, not a finding. What the factorial
does support is that the corpus makes no measurable difference at high
learning rate, that the diverse arm beats both narrow seeds on tool
correctness at low learning rate (with the diverse cell run once), and
that the completed narrow/low-LR run is the weakest arm on every axis
except the 31-item soft-miss axis; it does not license a
single-dominant-ingredient story. It did change one reading
elsewhere: the narrow/low-LR cell installs the call format, which is why
this paper no longer counts recipe B as a failure
(\S\ref{sec:results:factorial}). The two low-LR seed replicates that
would test Finding~1 were planned but not completed (L5); finishing both
(roughly 35 T4-hours) is the follow-up that bears most directly on it.

\paragraph{F4: Composition ablation at fixed scale (requires compute).}
Train 2--3 small models ($\approx$130--260M) identical except for realized
code fraction (e.g.\ 10\% / 35\% / 65\%), each with the same 0.1\% in-mix
SFT signal, and read out F1's trajectory probes---now a validated
instrument (\S\ref{sec:results:trajectory})---plus the strict
diagnostics. A
monotone relationship between code fraction and default tool-call emergence
at matched scale and token count would establish the compositional claim
causally; a variant that continue-pretrains~\cite{ibrahim2024simple} the 1B
on the code-heavy mix before SFT would test whether late composition can
substitute for early.

\section{Limitations}
\label{sec:limitations}

This series' policy is to own limitations plainly. The ones below are
material, and several would be disqualifying for stronger claims than the
ones we make.

\paragraph{L1: The 600M run is permanently frozen at 64\% of schedule.}
Training halted at step 154{,}000 of a planned 240{,}000
($\approx$10B of 15.7B tokens) and can never be resumed: the training VM and
the cloud account that held it no longer exist. The model never received its
cosine anneal and stopped at a still-elevated learning rate, which likely
caps its quality below what the recipe intended. Every 600M number in this
paper describes an incomplete artifact. For the paper's comparative claim
this biases \emph{against} us finding the effect (the handicapped model
shows default emergence anyway), but as a model release the 600M must be
understood as a frozen snapshot, not a finished product.

\paragraph{L2: No conversational capability, by construction---on either
side.}
The 600M had no dedicated conversational-SFT phase (unlike Nano and Base,
which had a three-phase OASST-style curriculum). It does not answer direct
simple questions: asked the capital of Peru, three fruits, $8+5$, or a
definition of phishing, it never produced a direct concluding answer in our
battery (\S\ref{sec:results:sanity}), though its Spanish remained fluent and
on-topic. Likewise, the \emph{repaired} 1B's fix is capability-specific: on
the one conversational item of the original eight-prompt battery, the
historical run correctly suppressed the tool call but returned an empty
answer, and the 238-prompt battery checks no-call answers only for being
non-degenerate, not for being correct. Nothing in this paper
claims a usable assistant on either side; the claims are about structured
tool-call emission only.

\paragraph{L3: The matched pair is not a controlled experiment.}
\label{sec:limitations:confounds}
The pair is matched in architecture family, tokenizer, and token layout, but
parameter count (1.68$\times$), total pretraining tokens
($\approx$10B vs.\ $\approx$65B), curriculum shape (single-phase vs.\
multi-phase), and training hardware all differ, and composition is
confounded with each. Our defense is directional: every confound is a
handicap for the 600M or an advantage for the 1B (fewer parameters, fewer
tokens, no anneal, no dedicated SFT), so the conventional explanations
predict the \emph{opposite} default. That makes composition the leading
surviving hypothesis for the default asymmetry, not a demonstrated cause;
the ablation that would demonstrate it (F4) has not been run.

\paragraph{L4: All strict diagnostics are small-$n$.}
The verbatim check used six examples (four in the repaired-1B evaluation
environment), all from \texttt{tool\_sft\_mini\_v1.jsonl}; the
generalization battery used eight novel prompts, constructed by us rather
than sampled from an external distribution. We attach no statistical claims
beyond the qualitative one: capabilities that appear in (nearly) every
trial for one configuration and never for another, under greedy decoding on
maximally favorable prompts, differ categorically. Whether the repaired
1B's 8/8 holds at hundreds of prompts, under adversarial phrasing, or on
multi-call sequences is untested (F2). The sharpest case of this is the
\emph{suppression} condition: it is a single pinned prompt, $n{=}1$ per
checkpoint, and it now carries a claim---that the 600M's default is an
unconditional base rate rather than a context gate
(\S\ref{sec:results:trajectory}, \S\ref{sec:discussion:mechanism}). What
that $n{=}1$ can support is a negative: no gate strong enough to show on
one clearly-forbidding prompt, replicated as a binary generation outcome
at 15 of 15 independent checkpoints. It cannot rule out a partial, noisy,
or prompt-specific gate, and it says nothing about how the 600M behaves
when a preamble discourages tool use less explicitly. F2 has since run
(\S\ref{sec:results:generalization}), lifting the verbatim check to
$n{=}269$ and the battery to $n{=}238$ including 72 suppression prompts,
and it upholds the base-rate correction at power. The preamble-strength
sweep remains untested, so the final clause of this limitation stands;
L11 records what the scaled battery itself cannot support.

\paragraph{L5: The repair is one recipe; its ingredients are separated
only descriptively.}
Recipe C bundled four simultaneous changes (corpus diversity, 5$\times$ LR,
unfrozen embeddings, a different warm-start point). Two of the four are now eliminated, neither by a
designed ablation: the warm-start point, because the recipe was run twice
from different parents with the same strict outcome
(\S\ref{sec:results:repair}), and the unfrozen embeddings, because direct
measurement shows the table was numerically inert across the run
(\S\ref{sec:results:mech}). Both eliminations are weaker than the
two-armed contrasts they stand in for: the two parents are siblings sharing
a frozen embedding table rather than independent draws, and an update that
underflows bf16 storage can still accumulate in fp32 optimizer state, so
``inert at this learning rate for this many steps'' is not ``inert.''
Corpus diversity and learning rate are no longer confounded with each
other: F3, the factorial that separates them, has now been run and is
reported in \S\ref{sec:results:factorial}. What it supports is narrower
than what it was designed to support, and three qualifications travel with
it. \emph{(i) Different numerical regime.} The five arms ran on a T4 in
fp16 with an 8-bit optimizer because Turing has no bf16, whereas the
recipe-C reproduction that supplies every other repaired-1B number here ran
in bf16 on an A10G. The arms are comparable to each other and to nothing
else in this paper; in particular they do not re-measure recipe C, and the
precision change touches exactly the quantity \S\ref{sec:results:mech}'s
underflow analysis turns on. \emph{(ii) One completed run per cell in three of four
cells.} Only the diverse/high-LR corner was replicated to completion.
Replicates of the two low-LR cells, the ones behind Finding~1, were
planned: the diverse/low-LR replicate was never started, and the
narrow/low-LR replicate was interrupted at step 880 of 2{,}202. Evaluated
at that step against the two original low-LR arms at the same step, the
second narrow seed suppresses as well as the diverse arm ($0.569$
vs.\ $0.556$) and far better than the first narrow seed ($0.389$; 13 items
to 0), so the suppression half of Finding~1 is not replicated and is held
as a hypothesis; the tool-correctness half holds against both narrow seeds,
with the diverse cell still run once (\S\ref{sec:results:factorial}). This
is a partial check: one extra seed, of one cell, at 40\% of training.
Finishing both low-LR replicates would take about 35 T4-hours and has not
been done. The high-LR replicated cell's two seeds,
while identical in aggregate headline pass, disagree on six individual
items and differ by $0.024$ on tool correctness---comparable to the
learning-rate effect measured inside the diverse corpus. Every interval we
report on that factorial is a paired bootstrap over \emph{items} and
carries no run-level variance, so the difference-in-differences that
suggests an LR$\times$corpus interaction describes five checkpoints rather
than estimating a population quantity. \emph{(iii) Correction matters.} The
factorial runs 36 tests; four survive Benjamini--Hochberg at $q{=}0.05$,
and two of those four (suppression pass and over-triggering at low LR) are
complements of one behavior rather than independent evidence. The claims we
draw are accordingly limited to: at high learning rate the corpus makes
no measurable difference; at low learning rate the diverse arm beats both
narrow seeds on tool correctness, while its large suppression advantage
over the completed narrow run is matched by seed-to-seed variation and is
not claimed; and the completed narrow/low-LR run is the weakest arm on
headline pass, tool correctness and suppression---though not on soft miss,
where it is nominally best, an axis scored on only 31 items and on which no
contrast in the factorial approaches significance. We do not claim the
factorial identifies a dominant ingredient. Separately, recipe B, the
narrow retune once counted as a second failed remediation, was never
measured with the strict instruments (\S\ref{sec:results:repair}), and no
claim in this paper now rests on it. The formal B4 number for the
repaired checkpoint is 8/8 tool-call emission, recovered only after fixing
a decode bug in the harness itself that initially reported 0/8 against
the same well-formed responses the strict diagnostic already scored 8/8
(\S\ref{sec:discussion:practical}); B1 (CVE keyword recall) remained
uninformative for an unrelated reason---the extracted ``keywords'' are the
corpus record's own schema labels, not content a paraphrased answer would
ever reproduce---and we do not report a corrected B1 here. The strict
diagnostics remain the evidence base.

\paragraph{L6: The mechanistic claim is localization by elimination, not
circuit analysis.}
``Routing, not representation'' rests on drift statistics between two
checkpoints under one recipe: the target row did not move while attention
weights did. This falsifies the input-representation hypothesis \emph{for
this repair}; it does not identify the circuit that changed, exclude
distributed small changes elsewhere in the embedding table from mattering,
or establish that the 600M's default operates through the same pathway the
repair built. One further caveat is specific to how the row's stillness
arose. \S\ref{sec:results:mech} attributes it to the bf16 quantization
floor---the per-step update is below the precision the weights are stored
in---rather than to a decision the training made. That strengthens the
negative claim (the ingredient cannot have caused the repair, because it
never took effect) while weakening any positive reading: it does not show
that a \emph{moved} embedding would have been useless, only that this run
never tested the question. The direct test is the frozen-vs-unfrozen
contrast at a learning rate high enough to clear the quantization floor,
which no run in this paper performs.

\paragraph{L7: All harness numbers are single-seed, over an export path
with history.}
B1--B5 scores are single-seed, as flagged throughout the series, and were
produced over the GGUF/Ollama path, in which this series has previously
found and fixed a RoPE export bug; NoPE-interleaved architectures have no
native \texttt{llama.cpp} support and export via a per-layer workaround.
The strict diagnostics deliberately bypass this path, but the B1--B5 table
inherits its risks.

\paragraph{L8: The 1B leads on B3, and we have not audited B3.}
On the lenient harness the 1B beats the 600M on command generation (0.24
vs.\ 0.17). Given that we discredit the same harness's B4 for the 1B, the
symmetric possibility---that B3's lenient matching also misestimates one or
both models, in either direction---is open. We report the number rather
than cherry-picking around it.

\paragraph{L9: Composition fractions are estimates.}
The realized 65/34/0.7/0.1 breakdown of \S\ref{sec:training:mix} is
reconstructed from bin sizes and sampler behavior, not from an exact audit
of the consumed token stream (the raw bins survive only partially). The
configured weights are exact; the realized fractions should be read as
good-faith estimates with uncertainty of a few percentage points, which
does not affect the qualitative code-heavy vs.\ web-heavy contrast.

\paragraph{L10: The trajectory sweep has three specific blind spots.}
\label{sec:limitations:trajectory}
The curves of \S\ref{sec:results:trajectory} carry limits beyond the
small-$n$ one of L4. \emph{(a) The rising side of the 1B's prior is
permanently unobservable.} Phase-2 step 36K is the earliest archived
checkpoint of the whole 1B run---phase 1 archived none---so we can measure
the decay but not the peak, the installation point, or whether 36K was
already well past its maximum. Every statement we make about the erasure
is bounded below by that checkpoint; ``nearly six orders of magnitude'' is
what we observed, not necessarily the full drop. \emph{(b) The probe set
is contaminated with respect to the 1B's phase-1 data.} All three
well-formed emissions at 36K are byte-exact reproductions of their
training completions with zero generalization, which places the pinned
examples from \texttt{tool\_sft\_mini\_v1.jsonl} at or very near phase-1
training text. We verified the contamination does not come from phase 2
(no \texttt{tool\_sft} bucket appears in any phase-2 block mixture), but
the 36K value must be read as a memorized prior washing out, not as a
capability being lost; a clean version of this measurement needs held-out
probe examples, which the 600M side would also benefit from. \emph{(c) The
B4 series paired against the probe is not the same harness invocation as
Table~\ref{tab:b15}'s.} It comes from the project's phase-3
checkpoint-tracking record, whose values for this lineage top out near
0.085, while Table~\ref{tab:b15} reproduces a recorded 0.650 for the same
model. We use the tracking series only for within-series comparison and
never as an absolute score, and we note the discrepancy itself as further
reason to distrust B4 for this model rather than as something we have
resolved. Additionally, the specific checkpoints anchoring the
dissociation (the 14K B4 peak, the 16K floor) are flagged in the project's
own record as single-seed volatility not to be over-interpreted
individually; the probe side is greedy and deterministic, but the
harness side of that comparison is exactly the noisy one.

\paragraph{L11: The generalization battery is ours, and four of its
design choices bound what it can show.}
\label{sec:limitations:battery}
\emph{(a) A scoring bug was found and fixed after the first analysis, and
it mattered.} The argument check was written as a disjunction whose second
clause subsumed the first, so a required key carrying \emph{any} non-empty
value passed---including a memorized corpus entity supplied in place of the
novel one the prompt asked about. Re-scoring the same saved generations
under value equality moved the fixed-schema pass rate from $0.726$ to
$0.474$ for the 600M and from $0.652$ to $0.541$ for the repaired 1B. The
corrected numbers are the ones reported throughout
\S\ref{sec:results:generalization}; we record the episode because the bug
flattered the 600M roughly twice as much as the 1B and, uncorrected,
produced a spurious tie between the two models. We report it rather than
silently shipping the fix because it is a reminder that the scorer is an
instrument with its own failure modes, exactly like the B4 harness this
paper spends \S\ref{sec:results:harness} discrediting.

\emph{(b) 18 of the 72 suppression prompts are duplicates.} The pure and
near-miss topic pools hold 33 and 20 items respectively and were sampled
with replacement to reach the battery's 30\% suppression target. The
duplication is declared in the generator rather than hidden, and it does
not affect the call-eliciting families (verified unique by Cartesian
sampling without replacement), but it means the effective $n$ on the
suppression axis is 53 distinct prompts, not 72, and the intervals we
report there are correspondingly optimistic.

\emph{(c) The \texttt{anchor} subset is reworded, not original.} The eight
prompts of \S\ref{sec:results:repair} were rebuilt from the paper's own
glosses when the battery was constructed, preserving the intent and the
expected tool but not the surface form (the original clipped
\texttt{``busca CVEs relacionados con log4j''} became
\texttt{``Buscá CVEs relacionados con log4j.''}, and so on). The literal
originals do survive in the probe script, so this was a normalization into
the battery's question format rather than a forced reconstruction from
memory---but the items scored here are the reworded ones, and
\S\ref{sec:results:generalization} independently finds register to be a
variable both models are sensitive to. Their agreement with the original
result (6/7 vs.\ 7/7) is therefore weak corroboration, not replication, and
the items are flagged \texttt{anchor\_reconstructed} in the battery
file.

\emph{(d) The soft-miss axis exists only for \texttt{bash\_exec}.} The
other four tools take fixed schemas, so there is no analogue of ``right
tool, wrong specific command'': once the tool and the entity are right
there is nothing further to get wrong. The soft-miss rates of
Table~\ref{tab:genaxes} therefore rest on 30--31 items each and describe
one tool's behavior, not a general property of either model. Relatedly,
\texttt{bash\_exec} is held to a strictly harder standard than the other
four families---tool, shape, \emph{and} command-family match, against tool,
shape, and entity---so the per-family ordering in
Table~\ref{tab:genfamily} should not be read as a clean difficulty ranking
across tools.

\paragraph{L12: The repair's original checkpoint and its parent are lost;
every repaired-1B number here comes from a reproduction.}
\label{sec:limitations:lineage}
The checkpoint produced by the historical recipe-C run, and the warm-start
parent it was compared against in the embedding-drift check, were both
destroyed when the infrastructure holding them was retired---a rented pod
that no longer exists and a cloud resource group whose sponsoring credit
expired. Neither survives in any archive we can reach. A third of the
recipe's corpus was also lost and was regenerated from its original source
at a pinned revision, reproducing the training loader's example count and
step total exactly. What this paper reports for the repaired 1B is
therefore a \emph{re-derivation}: the same corpus, learning rate and step
count, warm-started from a surviving sibling of the lost parent, gated
before launch on a CPU check confirming the substitute exhibits the same
pathology the repair targets. The historical figures are retained in
Tables~\ref{tab:verbatim} and~\ref{tab:recipes} for comparison and are
load-bearing for nothing. Four specific residues of this episode are worth
declaring.

\emph{(a) The substitute is not bit-identical to the pure phase-3
backbone, and the project's own notes said otherwise.} Phase 4a freezes the
language model, so an internal record described the substitute's LLM
weights as ``identical to phase 3's.'' Direct comparison refutes that: all
201 language-model tensors differ. The differences are minute---max
absolute difference $7.8{\times}10^{-3}$ across the whole model, cosine
similarity per tensor no lower than $0.999966$---but their \emph{shape} is
the informative part. Essentially no entries are bit-identical, which is
the opposite of the signature that low-learning-rate training leaves in
bf16 (most entries exactly unchanged plus a moving tail, as in
\S\ref{sec:results:mech}). A uniform, tiny, everywhere-nonzero perturbation
is what a precision round-trip in an export or conversion step looks like,
not what training looks like. We treat the substitute as a sound stand-in
on that basis---the distance to phase 3 comes from a source other than
optimization---while retracting the ``identical'' claim, which was never
verified.

\emph{(b) A candidate for the lost child was re-tested and ruled out.} One
archived checkpoint carried a name suggesting it was the repaired model.
Replaying the Level-3 verbatim diagnostic on it yields 1/6 well-formed:
the single correct suppression, with the other five producing bare JSON
without the \texttt{<|tool\_call|>} wrapper, prose, or empty completions.
Its recorded step count (2{,}002) does not match the recipe's 2{,}202
either. It is a different and substantially weaker artifact---plausibly an
incomplete or failed run---and the search for the original ended
unsuccessfully.

\emph{(c) The norm that was used to rule that candidate out does not mean
what the record assumed.} The candidate had been set aside earlier on the
grounds that its embedding-table norm read $630.13$ rather than the
$\approx$$635.2$ logged during the repair run. Both numbers are real
measurements of the same tensor by different code paths. On the version of
the framework in use, \texttt{torch.norm} and
\texttt{torch.linalg.vector\_norm} return $630.134705$ for this
$32{,}768{\times}2048$ bf16 table while an explicit
$\sqrt{\sum w^2}$ returns $635.1627$ in both fp32 and fp64---the explicit
computation being the accurate one, and the library reduction being in
error by 0.8\% at this tensor size. The ``post-repair signature'' and the
``pre-merge value'' were never evidence about which checkpoint was which.
This is the fourth instrument artifact this paper has had to disentangle
from a substantive finding, after B4's keyword scorer
(\S\ref{sec:results:harness}), the special-token decode bug
(\S\ref{sec:discussion:practical}), and the battery's own argument scorer
(L11a); we record it in that spirit. It also corroborates
\S\ref{sec:results:mech} from a second direction: under the accurate
reduction the table's norm moves from $635.162769$ to $635.163016$ across
the entire repair, a relative change of $4{\times}10^{-7}$.

\emph{(d) One thing remains genuinely unresolved.} No logs survive from the
historical run or from the job that produced its parent, so we cannot
establish which reduction path produced the $635.2$ figure recorded for
that parent at the time, and therefore cannot fully exclude that the
historical lineage differed from the reproduced one in some way the
pathology gate would not have caught. The reproduction agreeing with the
historical numbers to within reproduction noise on every instrument
(Tables~\ref{tab:verbatim},~\ref{tab:recipes}, and
\S\ref{sec:results:mech}) is evidence against that, not proof. Readers
should treat the historical row as an uncorroborated record and the
reproduced row as the artifact.

\section{Conclusion}
\label{sec:conclusion}

We followed a benchmark anomaly through a complete diagnose-and-repair arc
on a matched-architecture pair of small security language models. A lenient
keyword harness scored \model (661.6M, code-heavy single-phase bootstrap,
frozen at 64\% of schedule, no dedicated SFT) and \onebee (1{,}109M,
web-heavy multi-phase curriculum with a dedicated $\approx$6B-token
tool-SFT phase) within 0.01 of each other on tool use. Strict diagnostics
showed the numbers described opposite realities: the 600M emits well-formed,
generalizing \texttt{<|tool\_call|>} JSON by default (6/6), while the 1B
never produced the structure at any pre-remediation checkpoint (0/4--6),
with a first-token prior of $10^{-4}$--$10^{-5}$---making the 1B's harness
score a documented false positive of a failure class this series has now
observed twice. The diagnosis then paid for itself: where a
$\approx$6B-token dedicated SFT phase had failed, a diagnosis-informed recipe (diverse assistant-mode corpus,
5$\times$ LR, $\approx$3.3 GPU-hours) repaired the 1B to genuine,
generalizing five-way tool selection with novel-argument extraction
(8/8)---and an embedding-drift check located the repair in the network's
routing, not in the trigger token's representation, which had been
well-placed all along. That check went on to answer a question we had
budgeted a follow-up run for: the ``unfrozen'' embedding table of the
successful recipe turns out to be 97.7\% bit-identical to its parent, every
step's update having fallen below the precision the weights are stored in,
so the recipe already ran with embeddings effectively frozen.

The synthesis refines this series' central claim. Bootstrap corpus
composition does not decide whether small models \emph{can} call tools; it
decides whether the capability arrives \emph{by default} or must be
engineered in afterward---and SFT token volume is uninformative about which
intervention will work, since three orders of magnitude more tokens failed
where a correctly shaped $10^7$-token recipe succeeded. Alongside the
thesis we offer the instrument that made every step of the arc possible: a
minutes-of-CPU diagnostic ladder---verbatim reproduction with a
generalization criterion, first-token probes, a novel-prompt battery, and
an embedding-drift check---that caught the false positive, localized the
failure, verified the repair as more than memorization, and falsified the
obvious mechanism story.

The comparison retains acknowledged confounds, the diagnostics are
small-$n$, the original repair artifact was lost and what we report is a
reproduction of it, and neither model has been shown to answer a plain question usefully; the
follow-ups that would close each gap are specified. The one that mattered
most, a $2{\times}2$ factorial over the repair's two remaining ingredients,
has since been run, in a changed numerical regime and with one completed
run in three of its four cells, so we report it as a description of five
checkpoints rather than a causal estimate. At high learning rate the corpus
made no measurable difference. At low learning rate the diverse arm
suppressed calls far better than the narrow one, but a second seed of the
narrow cell, checked at 40\% of training, suppressed as well as the
diverse arm; that contrast is therefore a hypothesis, not a finding, until
the two low-LR replicates are finished.
Every cell of it, including the narrow low-LR corner, installs the call
format. A dedicated tool SFT of the 600M, with or without data distilled
from a larger model, raises its call-side pass rate but gives it no gate;
the distilled data has no measurable effect. For practitioners
the actionable summary is two sentences now: put the output format in the
pretraining mix, and verify tool use with a grader that can tell a tool
call from a sentence about one. And when a capability seems missing,
diagnose before you conclude---the difference between ``impossible'' and a
few dollars of GPU time was a measurement.

\paragraph{Artifact availability.} The following are public on the
Hugging Face Hub at the time of writing: the 600M pretraining checkpoints,
including the step-154K snapshot studied here
(\texttt{jsantillana/vectrayx-600m-checkpoints}); the shared tokenizer
(\texttt{jsantillana/vectrayx-600m-tokenizer}); and, in
\texttt{jsantillana/vectrayx-vision-1b-checks}, the reproduced repair
checkpoint with its intermediate steps (\texttt{text\_diag\_repro\_v0}),
its warm-start parent (\texttt{phase4a\_v0\_real\_llm}), and the final
checkpoints of all five factorial arms (\texttt{f3\_factorial\_v0}). The
historical repair checkpoint and its parent are lost (L12), and the
checkpoints of the 600M SFT ablation are not released.

\begin{acks}
The 600M pretraining run was carried out on a single NVIDIA L4 on Google
Cloud. The diagnostic experiments ran on commodity CPU hardware; the
historical repair run used a rented A40, its reproduction an A10G, the
factorial an NVIDIA T4, and the 600M SFT ablation an L4. We thank the maintainers of \texttt{llama.cpp} and the Hugging
Face Hub, on which the export and checkpoint-archival pipeline of this series
depends.
\end{acks}

\bibliographystyle{ACM-Reference-Format}
\bibliography{references}

@article{santillana2026vectrayx,
  title={VectraYX-Nano: A 42M-Parameter Spanish Cybersecurity Language Model with Curriculum Learning and Native Tool Use},
  author={Santillana, Juan S.},
  journal={arXiv preprint arXiv:2605.13989},
  year={2026}
}

@article{santillana2026vision,
  title={VectraYX-Vision-1B: A Sub-2B Spanish/LATAM Cybersecurity Vision--Language Model with Structured Visual Reasoning and Native Tool Use},
  author={Santillana, Juan S.},
  journal={arXiv preprint arXiv:2608.08477},
  year={2026}
}

@article{schick2023toolformer,
  title={Toolformer: Language Models Can Teach Themselves to Use Tools},
  author={Schick, Timo and Dwivedi-Yu, Jane and Dess{\`\i}, Roberto and Raileanu, Roberta and Lomeli, Maria and Hambro, Eric and Zettlemoyer, Luke and Cancedda, Nicola and Scialom, Thomas},
  journal={Advances in Neural Information Processing Systems (NeurIPS)},
  year={2023}
}

@article{patil2023gorilla,
  title={Gorilla: Large Language Model Connected with Massive APIs},
  author={Patil, Shishir G and Zhang, Tianjun and Wang, Xin and Gonzalez, Joseph E},
  journal={arXiv preprint arXiv:2305.15334},
  year={2023}
}

@inproceedings{qin2023toolllm,
  title={ToolLLM: Facilitating Large Language Models to Master 16000+ Real-World APIs},
  author={Qin, Yujia and Liang, Shihao and Ye, Yining and Zhu, Kunlun and Yan, Lan and Lu, Yaxi and Lin, Yankai and Cong, Xin and Tang, Xiangru and Qian, Bill and others},
  booktitle={International Conference on Learning Representations (ICLR)},
  year={2024}
}

@misc{mcp2024,
  title={Model Context Protocol Specification},
  author={{Anthropic}},
  year={2024},
  howpublished={\url{https://modelcontextprotocol.io}}
}

@article{aryabumi2024code,
  title={To Code, or Not To Code? Exploring Impact of Code in Pre-training},
  author={Aryabumi, Viraat and Su, Yixuan and Ma, Raymond and Morisot, Adrien and Zhang, Ivan and Locatelli, Acyr and Fadaee, Marzieh and {\"U}st{\"u}n, Ahmet and Hooker, Sara},
  journal={arXiv preprint arXiv:2408.10914},
  year={2024}
}

@inproceedings{madaan2022codelm,
  title={Language Models of Code are Few-Shot Commonsense Learners},
  author={Madaan, Aman and Zhou, Shuyan and Alon, Uri and Yang, Yiming and Neubig, Graham},
  booktitle={Conference on Empirical Methods in Natural Language Processing (EMNLP)},
  year={2022}
}

@article{kocetkov2022stack,
  title={The Stack: 3 TB of Permissively Licensed Source Code},
  author={Kocetkov, Denis and Li, Raymond and Ben Allal, Loubna and Li, Jia and Mou, Chenghao and Mu{\~n}oz Ferrandis, Carlos and Jernite, Yacine and Mitchell, Margaret and Hughes, Sean and Wolf, Thomas and others},
  journal={Transactions on Machine Learning Research (TMLR)},
  year={2023}
}

@article{penedo2024fineweb,
  title={The FineWeb Datasets: Decanting the Web for the Finest Text Data at Scale},
  author={Penedo, Guilherme and Kydl{\'\i}{\v{c}}ek, Hynek and Ben Allal, Loubna and Lozhkov, Anton and Mitchell, Margaret and Raffel, Colin and Von Werra, Leandro and Wolf, Thomas},
  journal={Advances in Neural Information Processing Systems (NeurIPS)},
  year={2024}
}

@misc{nvd,
  title={National Vulnerability Database},
  author={{NIST}},
  howpublished={\url{https://nvd.nist.gov}},
  year={2024}
}

@article{hoffmann2022chinchilla,
  title={Training Compute-Optimal Large Language Models},
  author={Hoffmann, Jordan and Borgeaud, Sebastian and Mensch, Arthur and Buchatskaya, Elena and Cai, Trevor and Rutherford, Eliza and de Las Casas, Diego and Hendricks, Lisa Anne and Welbl, Johannes and Clark, Aidan and others},
  journal={Advances in Neural Information Processing Systems (NeurIPS)},
  year={2022}
}

@article{abdin2024phi3,
  title={Phi-3 Technical Report: A Highly Capable Language Model Locally on Your Phone},
  author={Abdin, Marah and Jacobs, Sam Ade and Awan, Ammar Ahmad and others},
  journal={arXiv preprint arXiv:2404.14219},
  year={2024}
}

@article{allal2025smollm2,
  title={SmolLM2: When Smol Goes Big --- Data-Centric Training of a Small Language Model},
  author={Ben Allal, Loubna and Lozhkov, Anton and Bakouch, Elie and Bl{\'a}zquez, Gabriel Mart{\'\i}n and Penedo, Guilherme and Tunstall, Lewis and Marafioti, Andr{\'e}s and Kydl{\'\i}{\v{c}}ek, Hynek and Lajar{\'\i}n, Agust{\'\i}n Piqueres and Srivastav, Vaibhav and others},
  journal={arXiv preprint arXiv:2502.02737},
  year={2025}
}

@article{ibrahim2024simple,
  title={Simple and Scalable Strategies to Continually Pre-train Large Language Models},
  author={Ibrahim, Adam and Th{\'e}rien, Benjamin and Gupta, Kshitij and Richter, Mats L and Anthony, Quentin and Lesort, Timoth{\'e}e and Belilovsky, Eugene and Rish, Irina},
  journal={Transactions on Machine Learning Research (TMLR)},
  year={2024}
}

@inproceedings{ainslie2023gqa,
  title={GQA: Training Generalized Multi-Query Transformer Models from Multi-Head Checkpoints},
  author={Ainslie, Joshua and Lee-Thorp, James and de Jong, Michiel and Zemlyanskiy, Yury and Lebr{\'o}n, Federico and Sanghai, Sumit},
  booktitle={Conference on Empirical Methods in Natural Language Processing (EMNLP)},
  year={2023}
}

@article{su2024roformer,
  title={RoFormer: Enhanced Transformer with Rotary Position Embedding},
  author={Su, Jianlin and Ahmed, Murtadha and Lu, Yu and Pan, Shengfeng and Bo, Wen and Liu, Yunfeng},
  journal={Neurocomputing},
  volume={568},
  year={2024}
}

@article{kazemnejad2023nope,
  title={The Impact of Positional Encoding on Length Generalization in Transformers},
  author={Kazemnejad, Amirhossein and Padhi, Inkit and Natesan Ramamurthy, Karthikeyan and Das, Payel and Reddy, Siva},
  journal={Advances in Neural Information Processing Systems (NeurIPS)},
  year={2023}
}

@inproceedings{shazeer2020glu,
  title={GLU Variants Improve Transformer},
  author={Shazeer, Noam},
  booktitle={arXiv preprint arXiv:2002.05202},
  year={2020}
}

@article{zhang2019rmsnorm,
  title={Root Mean Square Layer Normalization},
  author={Zhang, Biao and Sennrich, Rico},
  journal={Advances in Neural Information Processing Systems (NeurIPS)},
  year={2019}
}

@inproceedings{sennrich2016bpe,
  title={Neural Machine Translation of Rare Words with Subword Units},
  author={Sennrich, Rico and Haddow, Barry and Birch, Alexandra},
  booktitle={Annual Meeting of the Association for Computational Linguistics (ACL)},
  year={2016}
}

@misc{llamacpp,
  title={llama.cpp: LLM inference in C/C++},
  author={Gerganov, Georgi and {llama.cpp contributors}},
  year={2023},
  howpublished={\url{https://github.com/ggerganov/llama.cpp}}
}

@misc{ggml2024gguf,
  title={GGUF File Format Specification},
  author={{ggml contributors}},
  year={2024},
  howpublished={\url{https://github.com/ggml-org/ggml/blob/master/docs/gguf.md}}
}

\end{document}